\documentclass[letterpaper]{article}

\usepackage[preprint]{aaai2027}

\usepackage{graphicx}
\usepackage{booktabs}
\usepackage{multirow}
\usepackage{makecell}
\usepackage{algorithm}
\usepackage{algpseudocode}
\usepackage{subcaption}
\usepackage{amsmath,amssymb}
\usepackage{xcolor}
\usepackage{enumitem}
\usepackage[hyphens]{url}
\usepackage{caption}
\usepackage{natbib}
\usepackage[capitalize]{cleveref}
\usepackage{nameref}
\newcommand{\secref}[1]{\emph{\nameref{#1}}}
\newcommand{\eg}{e.g.}

\newcommand{\cf}{cf.}

\newcommand{\vs}{vs.}

\providecommand{\keywords}[1]{}
\providecommand{\institute}[1]{}

\title{Seeking Physics in Diffusion Noise}

\author{
    Chujun Tang\textsuperscript{\rm 1,2}\thanks{Work done while Chujun Tang was a research intern at The Hong Kong University of Science and Technology (Guangzhou).},
    Lei Zhong\textsuperscript{\rm 3},
    Fangqiang Ding\textsuperscript{\rm 1}\thanks{Corresponding author: fangqiangd@hkust-gz.edu.cn.}
}
\affiliations{
    \textsuperscript{\rm 1}The Hong Kong University of Science and Technology (Guangzhou)\quad
    \textsuperscript{\rm 2}Brown University\quad
    \textsuperscript{\rm 3}University of Edinburgh\\
    \url{https://thegreatestcj.github.io/Physics-in-Noise/}
}
\begin{document}

\maketitle

\begin{abstract}
Do video diffusion models encode signals predictive of physical plausibility?
We probe intermediate denoising representations of pretrained Diffusion Transformers (DiTs) and find that physically plausible and implausible videos are partially separable in mid-layer feature space, even at high noise levels.
Within-source and perceptual-quality controls suggest that this signal is not fully explained by generator identity or generic visual quality.
We distill the signal into a lightweight, backbone-specific physics verifier trained on frozen features and use it in two complementary inference-time mechanisms under a fixed multi-trajectory budget:
progressive trajectory selection, which scores trajectories at intermediate checkpoints and prunes weak candidates early, and reward-gradient guidance, which steers surviving trajectories by backpropagating through only the first few DiT blocks.
Experiments on PhyGenBench and Physics-IQ across CogVideoX-2B/5B and Wan 2.1-14B show that progressive selection matches verifier-based Best-of-4 on CogVideoX-2B while reducing wall-clock inference time by 37\%, whereas reward-gradient guidance substantially improves physical consistency on CogVideoX-5B, all without fine-tuning the video generator.
\end{abstract}

\section{Introduction}
\label{sec:intro}

Video diffusion models~\cite{yang2025cogvideox,kong2024hunyuanvideo,polyak2024moviegen,brooks2024sora} now generate visually realistic videos from text.
However, systematic evaluations still reveal frequent violations of basic physical
commonsense, including inconsistent gravity, implausible collisions, and unrealistic
object dynamics~\cite{kang2025howfar,meng2025phygenbench,bansal2024videophy}. This exposes a persistent gap between perceptual realism
and physical plausibility.

Existing approaches generally fall into three categories. They either introduce
external physical guidance into a frozen generator through physics-conditioned
generation~\cite{liu2024physgen,le2025vlipp,yuan2025newtongen,
gillman2025forceprompting,liu2024physdreamer}, modify the generator through
physics-aware post-training~\cite{li2025pisaexperiments,wang2025wisa,
zhang2025thinkbefore,zhang2025videorepa,yuan2024instructvideo,
prabhudesai2024aligningvideo,wu2024boosting,liu2025videoreward}, or perform
post-hoc selection by generating $N$ candidates, scoring them with a
VLM~\cite{bai2025qwen25vl,openai2024gpt4o}, and retaining the best at a cost
that scales linearly with $N$. The first two categories typically rely on
domain-specific physical priors or substantial post-training, while the third
incurs inference cost linear in the number of candidates. Despite their differences, these approaches do not explicitly exploit the
generator's intermediate representations as an internal signal of physical
plausibility. We revisit this assumption
and ask: \emph{does a frozen video diffusion model already encode signals
predictive of physical plausibility in its intermediate representations?}

\begin{figure}[t]
\centering
\includegraphics[width=0.475\textwidth]{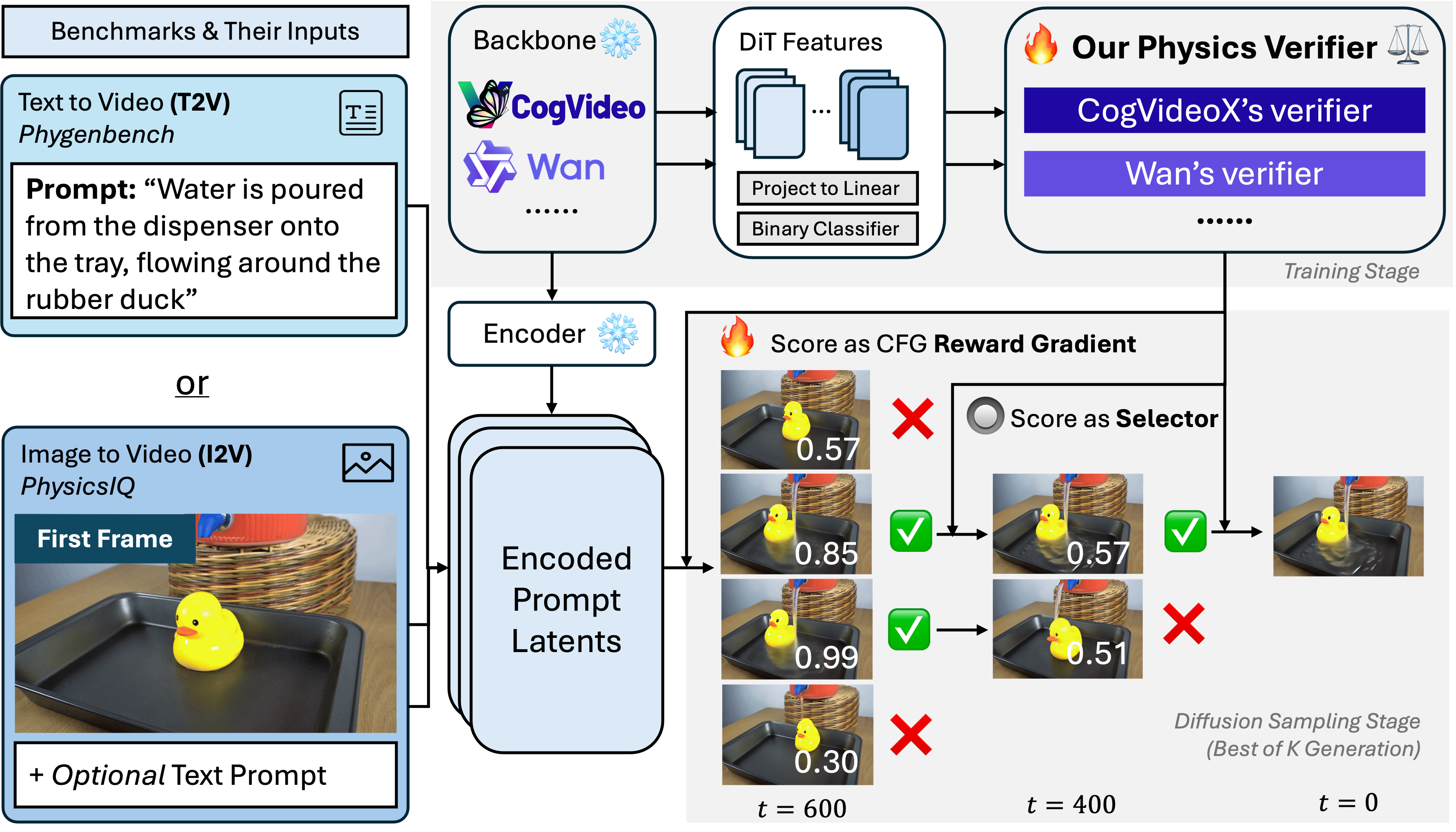}
\caption{\textbf{Overview of our physics-aware inference framework.} For both T2V and I2V generation, a lightweight
backbone-specific verifier scores intermediate DiT features. Its predictions
are used for reward-gradient guidance and progressive trajectory selection,
steering promising trajectories while pruning low-scoring candidates early.}
\label{fig:teaser}
\end{figure}

To answer this, we probe intermediate representations of frozen backbones
(CogVideoX-2B/5B~\cite{yang2025cogvideox} and
Wan~2.1-14B~\cite{wanteam2025wan}) along the denoising trajectory,
using human-annotated videos generated by diverse
models~\cite{bansal2024videophy}.
We find that physically plausible and implausible videos are partially separable
in mid-layer feature space, even at high noise levels.
A lightweight verifier with roughly 1M parameters, trained on these frozen
features, achieves an AUC of up to $0.68$, and the signal remains after
controlling for visual quality and generator identity.
These findings indicate that physical plausibility can be assessed before full
denoising and pixel-space decoding.

This early and differentiable signal motivates two complementary inference-time
mechanisms (\cf~Fig.~\ref{fig:teaser}).
Inspired by inference-time scaling, \emph{progressive trajectory selection}
runs $N$ trajectories in parallel and uses the verifier at intermediate
checkpoints to retain promising candidates while pruning low-scoring trajectories
early, thereby concentrating computation on trajectories more likely to yield
physically plausible videos.
Inspired by classifier guidance~\cite{dhariwal2021classifier,
ho2022classifier}, \emph{reward-gradient guidance} backpropagates the verifier
score to steer each surviving trajectory toward higher predicted physical
plausibility.
Because the score is computed from an intermediate block, the gradient traverses
only the preceding blocks, making guidance substantially cheaper than pixel-space
reward guidance~\cite{prabhudesai2024aligningvideo}, which backpropagates through
the full transformer and VAE decoder.
The two mechanisms are complementary: selection reallocates computation across
trajectories, whereas guidance improves each trajectory directly.
Both operate on a frozen backbone and require no model fine-tuning.

We evaluate our approach on PhyGenBench~\cite{meng2025phygenbench} and
Physics-IQ~\cite{motamed2025physicsiq} across
CogVideoX-2B/5B~\cite{yang2025cogvideox} and
Wan~2.1-14B~\cite{wanteam2025wan}.
On CogVideoX-2B, progressive trajectory selection matches Best-of-$N$
physical consistency while reducing wall-clock inference time by 37\%.
On CogVideoX-5B, where selection alone provides limited gains,
reward-gradient guidance substantially improves physical consistency.
These results demonstrate that intermediate diffusion features can support
both efficient search and direct trajectory steering without fine-tuning the
video generator.

Our contributions are:
\begin{itemize}[leftmargin=*,label=$\bullet$,itemsep=0pt,parsep=0pt,topsep=2pt]
    \item We show that physical plausibility is
    \emph{linearly decodable} from intermediate denoising features,
    remains detectable within individual generator sources, and is most
    salient in mid layers under substantial diffusion noise.

    \item We develop a lightweight physics verifier and two
    inference-time mechanisms: \emph{progressive trajectory selection},
    which prunes low-scoring candidates early, and
    \emph{reward-gradient guidance}, which directly steers trajectories.

    \item We validate the framework across multiple video diffusion
    backbones on physics-oriented T2V and I2V benchmarks, demonstrating
    improved physical plausibility and more efficient inference without
    fine-tuning the video generator.
\end{itemize}
\section{Related Work}
\label{sec:related_physics}

\noindent\textbf{Physical Understanding in Video Models.}
Strong generation quality does not imply physical competence:
Physics-IQ~\cite{motamed2025physicsiq} finds generative video models largely fail to
predict real-world dynamics, and VideoPhy~\cite{bansal2024videophy} and
PhyGenBench~\cite{meng2025phygenbench} report widespread commonsense violations
(see also~\cite{kang2025howfar}).
A complementary line probes \emph{internal representations}: intuitive-physics
signals emerge in V-JEPA-style self-supervised
predictors~\cite{garrido2025intuitive}; for generative models,
LikePhys~\cite{yuan2025likephys} reads physics preferences training-free from the
denoising likelihood, and concurrent work shows plausibility is linearly decodable
from intermediate diffusion features~\cite{esmati2026invisible,punzo2026layerwise}.
These establish that the signal exists but stop short of using it, with
\citet{esmati2026invisible} explicitly leaving probe-guided generation to future
work. Closest to our setting, WMReward~\cite{yuan2026wmreward} steers sampling with
an \emph{external} V-JEPA-2 reward.

\noindent\textbf{Physics-Aware Video Generation.}
One family conditions on explicit physical priors---motion
trajectories~\cite{liu2024physgen,le2025vlipp,geng2025motionprompting,zhang2025tora},
forces or goals~\cite{gillman2025forceprompting,gillman2026goalforce}, structured
dynamics and kinematic
constraints~\cite{yuan2025newtongen,wang2025physctrl,akkerman2025interdyn,romero2025kinemask},
or material properties~\cite{liu2024physdreamer}; \eg,
PhysGen~\cite{liu2024physgen} draws trajectories from rigid-body simulation and
NewtonGen~\cite{yuan2025newtongen} from neural Newtonian dynamics.
The achievable phenomena are bounded by the expressivity of the chosen prior.
A second family updates weights: supervised post-training on curated physics
data~\cite{li2025pisaexperiments,wang2025wisa,zhang2025thinkbefore,zhang2025videorepa},
or preference alignment from human, VLM, and learned-reward
feedback~\cite{yuan2024instructvideo,prabhudesai2024aligningvideo,cai2025phygdpo,wu2024boosting,liu2025videoreward}.
Both demand substantial compute and yield model-specific weights that do not
transfer across architectures.

\noindent\textbf{Inference-Time Scaling.}
Test-time search improves quality without touching weights.
In video, Video-T1~\cite{liu2025videot1} expands denoising trajectories in a tree,
DLBS~\cite{oshima2025dlbs} runs beam search with lookahead rollouts, and
EvoSearch~\cite{he2025evosearch} applies evolutionary search across diffusion and
flow generators.
The bottleneck is that reward models correlate poorly with final quality during
early denoising---\emph{verifier temporal coherence} in the terminology of
\citet{baraldi2025verifier}---and this is especially acute for physics: evaluators
such as PhyGenBench~\cite{meng2025phygenbench} and
VideoScore~\cite{he2024videoscore} are reliable only on decoded videos or require
costly rollouts, limiting their use as intermediate rewards.
\section{Probing Analysis}\label{sec:probing}

Post-hoc selection methods (\eg, Best-of-$N$) improve physical plausibility by
sampling multiple videos and choosing the best, but compute scales linearly with
$N$ \emph{full} denoising runs: without a reliable intermediate signal, selection
can only happen \emph{after} decoding~\cite{baraldi2025verifier}.
We therefore ask when and where physical knowledge emerges along denoising.

\begin{figure*}[!t]
    \centering
    \includegraphics[width=0.85\linewidth]{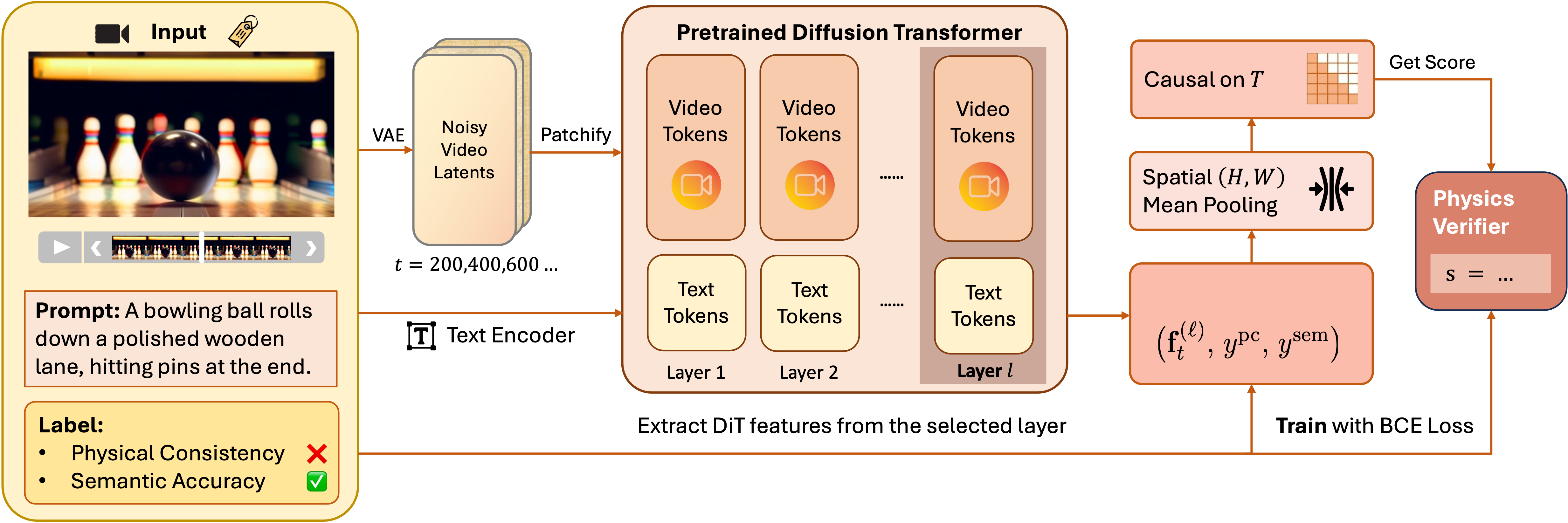}
    \caption{\textbf{Frozen-feature extraction and physics-verifier training.}
    Given a labeled video--prompt pair, we encode the video into VAE latents,
    add diffusion noise at timestep $t$, and pass the noisy latents through a
    frozen diffusion transformer. Hidden states from a selected layer $\ell$
    are extracted, with text tokens removed and video tokens spatially pooled
    to obtain per-frame features
    $\mathbf{f}^{(\ell)}_t \in \mathbb{R}^{F \times D}$.
    A lightweight verifier aggregates these features with causal temporal
    attention and predicts physical consistency and semantic accuracy, supervised
    by labels $y^{\mathrm{pc}}, y^{\mathrm{sem}} \in \{0,1\}$ using binary
    cross-entropy loss. The same frozen features are also used for the probing
    analysis.}
    \label{fig:pipeline_training}
\end{figure*}

\subsection{Setup}
\label{sec:feature_extraction}

We probe a frozen CogVideoX-2B~\cite{yang2025cogvideox} DiT on
VideoPhy~\cite{bansal2024videophy,bansal2025videophy2}, ${\sim}$4,500 videos from
seven text-to-video generators, each annotated for physical commonsense (PC) and semantic
accuracy (SA). We use the corresponding binary labels for
probing and verifier training, with PC positives defined as
$\mathrm{PC}\geq3$.
As illustrated in Fig.~\ref{fig:pipeline_training}, we encode each video into VAE
latents~\cite{rombach2022high}, add noise at $t \in \{200,400,600\}$ (larger $t$ is
noisier), run one forward pass, and take hidden states after block
$\ell \in \{5,10,15,20,25\}$; we drop text-conditioning tokens and spatially
mean-pool the video tokens into $\mathbf{f}^{(\ell)}_t \in \mathbb{R}^{F \times D}$,
with $F$ latent frames and hidden size $D$.
We flatten $\mathbf{f}^{(\ell)}_t$ over time, fit a logistic regression probe for PC,
and report mean AUC-ROC under 5-fold cross-validation (CV).
Running the \emph{same} probe on raw noised VAE latents isolates the DiT's
contribution: outperforming this baseline indicates that physics-related signal
becomes more linearly accessible \emph{through} denoising representations.

\subsection{Disentangling Source and Physics Signals}
\label{sec:source}

\begin{figure*}[!t]
\centering
\begin{subfigure}[t]{0.475\linewidth}
    \includegraphics[width=\linewidth]{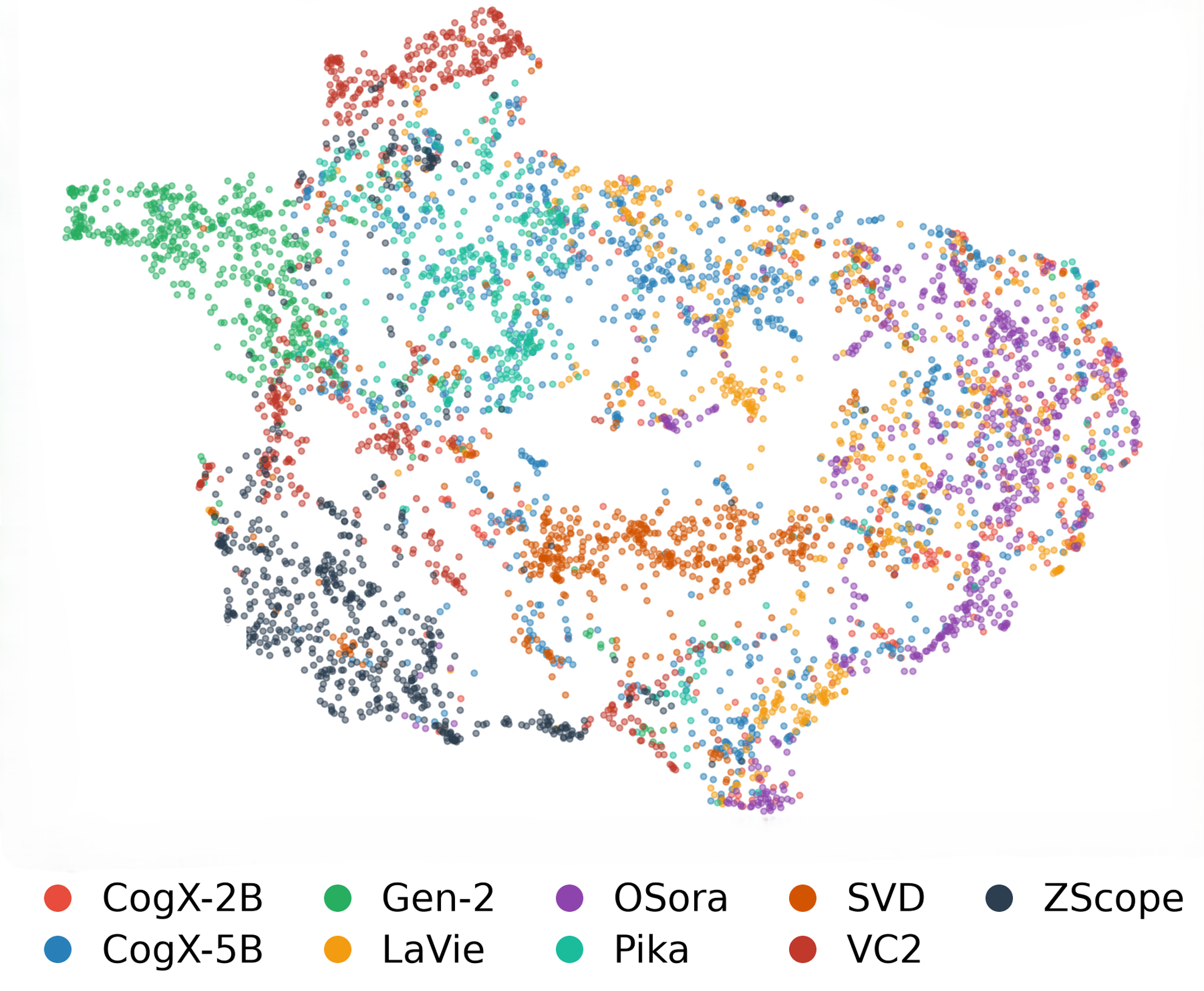}
    \caption{Feature space by generator}
    \label{fig:umap_source}
\end{subfigure}
\hfill
\begin{subfigure}[t]{0.475\linewidth}
    \includegraphics[width=\linewidth]{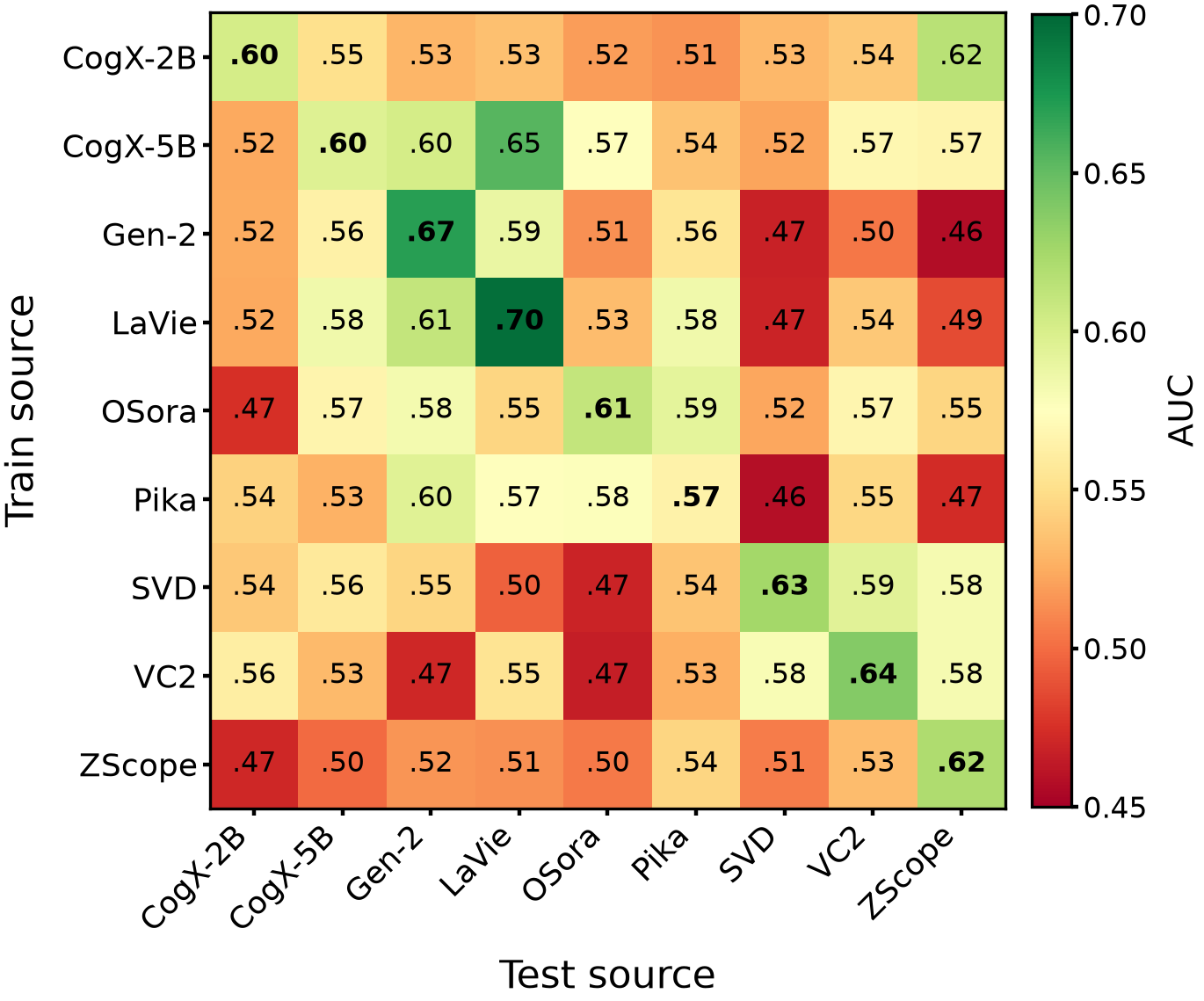}
    \caption{Cross-source probing (AUC)}
    \label{fig:source_cross}
\end{subfigure}
\caption{\textbf{Source structure in DiT features.} (a)~UMAP of CogVideoX-2B DiT features extracted after $\ell{=}10$ at $t{=}200$, colored by the original video generator, reveals strong source clustering. (b)~Cross-source probing AUC matrix: within-source evaluation (diagonal, 5-fold CV; mean 0.611) is substantially higher than cross-source transfer (off-diagonal mean 0.538).}
\label{fig:source_analysis}
\end{figure*}

Since VideoPhy aggregates seven generators with distinct styles and artifacts, the
\emph{video source} is a natural confounder.
Figure~\ref{fig:umap_source} confirms this: features cluster strongly by generator
identity, indicating that source-specific characteristics dominate the
representation space.
To control for it, we retrain probes \emph{within} each source (5-fold CV).
PC separability survives in every source (AUC 0.534--0.712,
Table~\ref{tab:within_source}), so DiT features encode PC-relevant information as a
secondary signal beneath the dominant source structure.
This motivates a \emph{matched-distribution} strategy that we train the verifier only on
videos from the generator used at inference.

\begin{table}[t]
\centering
\small
\setlength{\tabcolsep}{6pt}
\begin{tabular}{@{}lrc@{}}
\toprule
Source & \# Videos & AUC \\
\midrule
LaVie~\cite{wang2025lavie}          & 528 & 0.712 \\
VideoCrafter2~\cite{chen2024videocrafter2} & 468 & 0.631 \\
ZeroScope~\cite{cerspense2023zeroscope}    & 612 & 0.613 \\
Gen-2~\cite{runwaygen2}       & 477 & 0.612 \\
OpenSora~\cite{opensora2024}   & 610 & 0.597 \\
SVD~\cite{svd2023stability}         & 492 & 0.579 \\
Pika~\cite{li2025pika}              & 376 & 0.534 \\
\midrule
Overall & 3{,}563 & 0.652 \\
\bottomrule
\end{tabular}
\caption{Within-source linear probing AUC using
CogVideoX-2B~\cite{yang2025cogvideox} DiT features. We report results on
the seven non-CogVideoX sources in VideoPhy~\cite{bansal2024videophy}.}
\label{tab:within_source}
\end{table}

\subsection{Layer and Timestep Analysis}
\label{sec:layer_timestep}

Table~\ref{tab:probing_grid} reports probing AUC across layers and noise levels.
DiT features beat the VAE latent baseline in every configuration (up to $+0.101$ at
$\ell{=}10$, $t{=}600$), so the blocks amplify physics-relevant structure.
Middle layers carry the most signal ($\ell{=}10$: 0.638), while $\ell{=}5$ and
$\ell{=}25$ are weaker, consistent with vision transformers, whose middle layers
capture richer semantics~\cite{raghu2021vision,caron2021emerging}.
Finally, separability does not degrade with noise: at $\ell{=}10$, AUC at $t{=}600$
matches or exceeds $t{=}200$, so physics-related structure remains recoverable at
moderate noise.

\begin{table}[!t]
\centering
\resizebox{0.475\textwidth}{!}{%
\begin{tabular}{@{}lcccccc@{}}
\toprule
Timestep & VAE & $\ell{=}5$ & $\ell{=}10$ & $\ell{=}15$ & $\ell{=}20$ & $\ell{=}25$ \\
\midrule
$t{=}200$ & 0.539 & 0.569 & \textbf{0.606} & 0.604 & 0.574 & 0.579 \\
$t{=}400$ & 0.534 & 0.545 & 0.610 & 0.608 & \textbf{0.624} & 0.594 \\
$t{=}600$ & 0.537 & 0.565 & \textbf{0.638} & 0.595 & 0.583 & 0.585 \\
\bottomrule
\end{tabular}
}
\caption{Linear probing AUC (5-fold CV) on CogVideoX-2B DiT features. Every DiT
layer improves on VAE probing, by up to $+0.101$ ($\ell{=}10$, $t{=}600$). Best DiT
layer per row in \textbf{bold}.}
\label{tab:probing_grid}
\end{table}

\subsection{Control Analyses}
\label{sec:controls}

Two alternative explanations remain.
First, the signal might originate in the VAE encoder rather than the DiT; probing
raw noised latents across all settings shows DiT features consistently win, most
starkly on LaVie-sourced videos (0.730 \vs\ 0.537).
Second, physically plausible videos might simply score better under generic
evaluators. Using VQAScore~\cite{lin2024evaluating} as a coarse proxy, a probe
trained on VQAScore alone reaches only AUC 0.555, and residualizing the linearly
VQAScore-correlated component out of the features leaves probing AUC nearly
unchanged.
The signal is thus enhanced by the DiT and not reducible to perceptual quality.
\section{Method}
\label{sec:method}

\noindent\textbf{Problem.}
A frozen video diffusion model maps an initial noise
$\mathbf{z}_T$ and a prompt $\mathbf{c}$ to a video
$\hat{\mathbf{v}}$.
Sampling is stochastic, and physical plausibility can vary
substantially across generations from the same prompt, yet the
sampler provides no intermediate signal for identifying which
trajectory is more likely to obey physical constraints.
We ask how to improve the plausibility of the returned video under
a fixed initial trajectory budget, while leaving the backbone
weights unchanged.

\noindent\textbf{Overview.}
Our method has three components.
(i)~A lightweight \emph{physics verifier} predicts physical
plausibility from frozen DiT features extracted during denoising
(\secref{sec:physics_head}).
(ii)~At inference time, we denoise $N$ trajectories in parallel
and progressively prune low-scoring candidates, concentrating
computation on more promising trajectories
(\secref{sec:selection}).
(iii)~Between selection checkpoints, we steer each surviving
trajectory using the verifier gradient, injected in the same
additive form as classifier-free guidance
(\secref{sec:reward_guidance}).
The latter two mechanisms are complementary: selection
\emph{reallocates} computation across trajectories but does not
alter their evolution, so its performance is bounded by the best
trajectory induced by the $N$ initial noise samples; guidance
instead \emph{modifies} the trajectories themselves.
Selection requires no backpropagation: each checkpoint scores the active
trajectories with one conditional forward pass apiece, about 5\% of total
denoising cost, whereas guidance backpropagates only through the first
$\ell_{\mathrm{best}}$ DiT blocks.

\noindent\textbf{Notation.}
Let $\mathbf{z}_t$ denote the noisy latent at diffusion timestep
$t$, $\mathbf{c}$ the text embedding,
$\mathbf{h}_t^{(\ell)} \in \mathbb{R}^{S \times D}$ the
token-level hidden states after block $\ell$, and
$\mathbf{f}_t^{(\ell)} \in \mathbb{R}^{F \times D}$ the
spatially pooled features defined in
\secref{sec:feature_extraction}.
We use $\mathcal{T}$ and $\mathcal{L}$ to denote the candidate
diffusion timesteps and DiT blocks, respectively,
$\mathcal{C} \subseteq \mathcal{T}$ for selection checkpoints,
and $\mathcal{G} \subseteq \mathcal{T}$ for guidance steps.

\subsection{Physics Verifier}
\label{sec:physics_head}

Physical violations unfold over time and propagate to later frames, so we model
temporal dependencies with a single causal self-attention
layer~\cite{wang2021causal}: a triangular mask lets frame $i$ attend only to
$j \le i$, preventing the verifier from exploiting future frames when applied
mid-denoising.
For each $(t,\ell) \in \mathcal{T} \times \mathcal{L}$ we project
$\mathbf{f}_t^{(\ell)}$ to dimension $d$, add learnable positional embeddings
$\mathbf{p}_{1:F}$, and apply causal attention with a residual connection:
\begin{align}
  \tilde{\mathbf{f}}_t^{(\ell)}
  &= \mathrm{Proj}\bigl(\mathbf{f}_t^{(\ell)}\bigr) + \mathbf{p}_{1:F}
  \in \mathbb{R}^{F \times d},
  \label{eq:proj}\\
  \mathbf{y}_t^{(\ell)}
  &= \tilde{\mathbf{f}}_t^{(\ell)}
    + \mathrm{CausalAttn}\!\bigl(\mathrm{LN}(\tilde{\mathbf{f}}_t^{(\ell)})\bigr).
  \label{eq:causal}
\end{align}
The last-frame representation $(\mathbf{y}_t^{(\ell)})_{F}$ is therefore a
causally aggregated summary of the sequence; a LayerNorm and two-layer MLP map it
to a plausibility score
$s_t^{(\ell)} = \mathrm{Sigmoid}(\mathrm{MLP}(\mathrm{LN}((\mathbf{y}_t^{(\ell)})_{F}))) \in [0,1]$.
A second head with identical architecture predicts semantic accuracy.

\noindent\textbf{Training.}
We train one verifier per layer $\ell \in \mathcal{L}$ on
VideoPhy~\cite{bansal2024videophy}, applying forward diffusion at $t \in \mathcal{T}$
and extracting $\mathbf{f}_t^{(\ell)}$.
With targets $y^{\text{pc}}, y^{\text{sem}} \in \{0,1\}$ and head outputs
$s_{t,\text{pc}}^{(\ell)}, s_{t,\text{sem}}^{(\ell)}$, the loss is
\begin{equation}
\begin{split}
\mathcal{L}^{(\ell)} = \frac{1}{|\mathcal{T}|} \sum_{t \in \mathcal{T}}
\mathbb{E} \big[
  &\lambda_{\text{pc}}\,\text{WBCE}(y^{\text{pc}}, s_{t,\text{pc}}^{(\ell)}) \\
  &+ \lambda_{\text{sem}}\,\text{WBCE}(y^{\text{sem}}, s_{t,\text{sem}}^{(\ell)})
\big],
\end{split}
\end{equation}
where $\mathrm{WBCE}(y,s) = -\,w_y(y\log s + (1-y)\log(1-s))$, $w_y$ corrects class
imbalance, and $\lambda_{\text{pc}}, \lambda_{\text{sem}}$ balance the two terms.
The frozen backbone means one forward pass per video per timestep suffices; see
the \secref{sec:exp_setup} section for details.

\subsection{Progressive Trajectory Selection}
\label{sec:selection}

We pick the best layer $\ell_{\mathrm{best}}$ on validation, then denoise $N$
trajectories from independent noise samples $\{\mathbf{z}_T^{(i)}\}_{i=1}^{N}$,
eliminating implausible candidates along the way.
At each checkpoint $t \in \mathcal{C} \subseteq \mathcal{T}$ (ordered high to low
noise), the verifier scores all active trajectories from their layer-$\ell_{\mathrm{best}}$
features and retains the top fraction $\rho$, so the pool shrinks geometrically
(Algorithm~\ref{alg:selection}, Fig.~\ref{fig:teaser}).
Checkpoints are placed only in the moderate-noise regime where
\secref{sec:probing} found PC separability reliable, and we select on the PC score
alone: adding the semantic score did not help in preliminary experiments, likely
because classifier-free guidance already promotes semantic alignment.

Pixel-space verifiers are the usual bottleneck here: they lose temporal coherence
on noisy intermediate samples~\cite{baraldi2025verifier}, making early selection
unreliable.
Ours is trained directly on noisy DiT features at the same timesteps where
selection occurs, so it stays reliable even at $t{=}600$, enabling aggressive early
pruning where savings are largest.

\noindent\textbf{Computational cost.}
With $T_{\mathrm{steps}}$ DDIM steps, keep ratio $\rho$, and $K$ roughly evenly
spaced checkpoints, the expected number of DiT forward passes is
\begin{align}
  C_{\text{ours}}
  \approx T_{\mathrm{steps}} \cdot N \cdot \frac{1 - \rho^{K+1}}{(K{+}1)(1{-}\rho)}.
  \label{eq:cost}
\end{align}
With $N{=}4$, $K{=}2$, checkpoints at $t \in \{600,400\}$, and a 50-step schedule,
the pool shrinks $4 \rightarrow 2 \rightarrow 1$ ($20$ steps at $4$, $10$ at $2$,
$20$ at $1$), totaling ${\approx}120$ passes \vs\ $200$ for Best-of-4, an
${\approx}1.7\times$ speedup, with negligible verifier overhead.

\begin{algorithm}[t]
\caption{Progressive Trajectory Selection}
\label{alg:selection}
\begin{algorithmic}[1]
\Require Text embedding $\mathbf{c}$; number of trajectories $N$; checkpoints $\mathcal{C}$; keep ratio $\rho$; selected layer $\ell_{\mathrm{best}}$; physical verifier $f_\phi$; maximum diffusion timestep $T$
\Ensure Generated video $\hat{\mathbf{v}}$

\State Sample initial noises $\mathbf{z}_T^{(i)} \sim \mathcal{N}(\mathbf{0}, \mathbf{I})$ for $i=1,\dots,N$
\State Initialize active trajectory set $\mathcal{A} \gets \{1,\dots,N\}$

\For{$t = T, T-1, \dots, 1$}
    \For{$i \in \mathcal{A}$}
        \State $\mathbf{z}_{t-1}^{(i)}, \mathbf{h}_t^{(i)}
        \gets \text{DiTStep}(\mathbf{z}_t^{(i)}, t, \mathbf{c})$
        \State $\mathbf{f}_t^{(i)} \gets \text{PoolFeat}(\mathbf{h}_t^{(i)}; \ell_{\mathrm{best}})$
    \EndFor

    \If{$t \in \mathcal{C}$}
        \For{$i \in \mathcal{A}$}
            \State $s_t^{(i)} \gets f_\phi(\mathbf{f}_t^{(i)})$
        \EndFor
        \State Keep only the top $\lceil \rho |\mathcal{A}| \rceil$ trajectories ranked by $s_t^{(i)}$
    \EndIf
\EndFor

\State Let $i^*$ be the remaining trajectory in $\mathcal{A}$
\State \Return $\hat{\mathbf{v}} \gets \text{VaeDecode}(\mathbf{z}_0^{(i^*)})$
\end{algorithmic}
\end{algorithm}

\subsection{CFG-Style Reward Gradient Guidance}
\label{sec:reward_guidance}

Selection only reallocates compute: each trajectory evolves untouched, so quality
is capped by what the $N$ initial noise samples happen to offer.
We add a complementary \emph{generative} mechanism that steers every active
trajectory, injecting the verifier gradient in the same additive form as
classifier-free guidance (CFG):
\begin{equation}
  \hat{\boldsymbol{\epsilon}}_t
  = \boldsymbol{\epsilon}_\theta(\mathbf{z}_t, \varnothing)
  + w_{\mathrm{cfg}}\bigl(
      \boldsymbol{\epsilon}_\theta(\mathbf{z}_t, \mathbf{c})
      - \boldsymbol{\epsilon}_\theta(\mathbf{z}_t, \varnothing)
    \bigr).
  \label{eq:cfg}
\end{equation}
CFG ascends the log-likelihood of an \emph{implicit} classifier
$p(\mathbf{c} \mid \mathbf{z}_t)$~\cite{ho2022classifier}; our verifier is an
\emph{explicit} one for physical plausibility.
Writing it as a differentiable function of the latent,
$r_t(\mathbf{z}) = f_\phi(\mathrm{PoolFeat}(\mathbf{h}_t(\mathbf{z}); \ell_{\mathrm{best}})) \in [0,1]$,
we augment Eq.~\eqref{eq:cfg} with a matching term:
\begin{equation}
  \tilde{\boldsymbol{\epsilon}}_t
  = \hat{\boldsymbol{\epsilon}}_t
  - w_{\mathrm{phy}}\,\sigma_t\,
    \nabla_{\mathbf{z}_t} \log r_t(\mathbf{z}_t),
  \label{eq:reward_guidance}
\end{equation}
with noise scale $\sigma_t$ and strength $w_{\mathrm{phy}}$.
Since $r_t = \mathrm{Sigmoid}(u_t)$, we have
$\nabla \log r_t = (1 - r_t)\nabla u_t$: guidance damps automatically once the
verifier is confident and is strongest for borderline trajectories, precisely
those selection alone cannot rescue.

\noindent\textbf{Cost and schedule.}
Unlike pixel-space reward guidance, which backpropagates through the full backbone
and VAE decoder, our gradient flows only through blocks
$1,\dots,\ell_{\mathrm{best}}$ and the verifier.
With $\ell_{\mathrm{best}}{=}10$, one guidance step costs
${\approx}2\ell_{\mathrm{best}}/L$ extra forward-pass equivalents, a fraction
of one denoising step,
applied only at $\mathcal{G} \subseteq \mathcal{T}$.
Because the verifier is calibrated only at moderate noise
(\secref{sec:layer_timestep}), ascending it elsewhere risks reward hacking and
off-manifold drift; we therefore restrict guidance to the selection window
(\eg, $t \in [400,600]$), clip per-frame gradient norms, and keep
$w_{\mathrm{phy}} \ll w_{\mathrm{cfg}}$.
The two mechanisms compose inside Algorithm~\ref{alg:selection}: between
checkpoints every trajectory is \emph{improved}, and the checkpoint then
\emph{drops} those that failed to respond, so a reward-hacked trajectory that
drifts off-manifold is penalized at later, cleaner checkpoints.

\section{Experiments}
\label{sec:experiments}

\providecommand{\pending}{\textcolor{red}{\textbf{--}}}

\begin{table*}[!t]
\centering
\small
\setlength{\tabcolsep}{5pt}
\resizebox{0.95\textwidth}{!}{%
\begin{tabular}{@{}llccccccccc@{}}
\toprule
\textbf{Backbone} & \textbf{Method} & \textbf{Final} & \textbf{S1} & \textbf{S2} & \textbf{S3} & \textbf{Mech.} & \textbf{Opti.} & \textbf{Ther.} & \textbf{Mate.} & \textbf{Pairw.} \\
\midrule
\multirow{3}{*}{CogVideoX-2B~\cite{yang2025cogvideox}}
  & Base Model               & 0.370$^*$ & ---  & ---  & ---  & 0.38 & 0.43 & 0.34 & 0.39 & --- \\
  & Best-of-$N$            & 0.515 & 1.96 & 0.87 & \textbf{1.73} & 0.50 & \textbf{0.59} & 0.46 & 0.48 & --- \\
  & $+$ Selection (Ours)   & \textbf{0.515} & 1.98 & \textbf{0.91} & 1.69 & 0.49 & 0.58 & 0.47 & 0.49 & \textbf{66.1\%} \\
\midrule
\multirow{5}{*}{CogVideoX-5B~\cite{yang2025cogvideox}}
  & Base Model               & 0.363 & 1.54 & 0.58 & 1.21 & 0.283 & 0.493 & 0.322 & 0.308 & --- \\
  & $+$ Selection (Ours)   & 0.365 & 1.52 & 0.53 & 1.30 & 0.292 & 0.456 & 0.256 & 0.408 & 60.2\% \\
  & $+$ Reward Gradient    & 0.496 & \textbf{1.79} & \textbf{0.95} & \textbf{1.71} & \textbf{0.417} & \textbf{0.607} & \textbf{0.456} & \textbf{0.467} & \textbf{66.7\%} \\
  & DPO                    & 0.475 & 1.74 & 0.78 & 1.68 & 0.375 & 0.600 & 0.411 & \textbf{0.467} & 50.3\%$^\dagger$ \\
\midrule
\multirow{4}{*}{Wan~2.1-14B~\cite{wanteam2025wan}}
  & Base Model               & 0.569 & 2.05 & 1.28 & 1.79 & 0.525 & 0.740 & 0.489 & 0.458 & --- \\
  & $+$ Selection (Ours)   & \textbf{0.612} & \textbf{2.09} & \textbf{1.46} & \textbf{1.86} & \textbf{0.600} & \textbf{0.767} & \textbf{0.533} & \textbf{0.492} & --- \\
  & $+$ Reward Gradient    & 0.606 & 2.09 & 1.36 & 1.94 & 0.575 & 0.767 & 0.556 & 0.475 & 50.0\%$^\dagger$ \\
  & DPO                    & 0.558 & 2.08 & 1.36 & 1.53 & 0.600 & 0.727 & 0.433 & 0.400 & 50.3\%$^\dagger$ \\
\bottomrule
\end{tabular}
}
\caption{\textbf{Main results} on PhyGenBench across three backbones.
$^*$Official single-seed CogVideoX-2B result
from~\cite{meng2025phygenbench}.
Pairwise results compare each method against the corresponding
base model. For standard entries, we report the win rate among
non-tied comparisons.
$^\dagger$For these comparisons, the judge returns ties for nearly all
prompts (Wan reward gradient: 157/160; DPO: 153/160 on
CogVideoX-5B and 157/160 on Wan).
The displayed tie-adjusted rates are therefore provided only for
completeness and should not be interpreted as conventional win rates.}
\label{tab:cross_backbone}
\end{table*}

\subsection{Setup}
\label{sec:exp_setup}

\noindent\textbf{Backbones.}
We evaluate on three frozen video diffusion models: CogVideoX-2B~\cite{yang2025cogvideox} ($D{=}1920$, 30 blocks),
CogVideoX-5B~\cite{yang2025cogvideox} ($D{=}3072$, 42 blocks), and
Wan~2.1-14B~\cite{wanteam2025wan} ($D{=}5120$, 40 blocks).
Videos are generated at $480{\times}720$ (49 frames) with
classifier-free guidance scale 6.0.
For trajectory selection we initialize $N{=}4$ trajectories and select at
$t \in \{600,400\}$ with keep ratio $\rho{=}0.5$, reducing the pool
$4 \rightarrow 2 \rightarrow 1$.
Backbone-specific configuration (schedulers, CFG modes, feature hooks) is given in
the supplementary.

\paragraph{Verifiers.}
Following the matched-distribution principle, we train a separate verifier for
each target backbone using backbone-matched videos labeled for
physical commonsense (PC) and semantic accuracy (SA) under the
VideoPhy protocols~\cite{bansal2024videophy,bansal2025videophy2}. For each video, we
apply forward diffusion at $t\in\{200,400,600\}$ and extract frozen
features from layer $\ell=10$. This yields 343, 1,736, and 591
annotated videos for CogVideoX-2B, CogVideoX-5B, and Wan
2.1-14B, respectively. We split each dataset at the video level
into 85\% training and 15\% validation sets.

The verifier architecture is shared across backbones, while the
input projection adapts to their different hidden dimensions,
resulting in 0.55M--1.37M trainable parameters. We train with
AdamW, a learning rate of
$10^{-3}$, weight decay $0.01$, batch size 32, and OneCycleLR
with 10\% warmup. Class imbalance is handled using weighted BCE,
and training is stopped with patience 20 based on validation AUC.
Each verifier trains in under 10 minutes on a single A100 GPU.

\noindent\textbf{Baselines.}
\textbf{Base Model}: base model single-seed generation.
\textbf{Random Sel.}: identical drop schedule with random verifier scores given.
\textbf{Best-of-$N$}: run all $N{=}4$ trajectories to completion and select with verifier at $t{=}200$.
\textbf{$+$ Selection (Ours)}: progressive selection (\secref{sec:selection}).
\textbf{$+$ Reward Gradient (Ours)}: selection plus CFG-style gradient
guidance (\secref{sec:reward_guidance}).
\textbf{DPO}: Diffusion-DPO~\cite{wallace2024diffusiondpo} on LoRA adapters, trained on 2K preference pairs built from the same VideoPhy physics annotations. 
\textbf{LikePhys}~\cite{yuan2025likephys}: training-free likelihood-based physics
scorer, compared against our verifier on AUC only.

\noindent\textbf{Benchmarks.}
\textbf{PhyGenBench}~\cite{meng2025phygenbench}: 160 text prompts covering 27
physical laws in four categories (mechanics, optics, thermal, material), scored by
their PhyGenEval protocol, including VQAScore~\cite{lin2024evaluating} with
CLIP-FlanT5-XXL for single-frame phenomena (S1), GPT-4o~\cite{openai2024gpt4o}
multi-frame verification of event ordering (S2), and GPT-4o naturalness on a 0--3
scale (S3), aggregated per video.
\textbf{Physics-IQ}~\cite{motamed2025physicsiq}: conditions on the starting frames
of a real scene and scores the continuation against the actual future.
\textbf{Pairwise}: GPT-4o pairwise judgments against base model, order
randomized.

\subsection{Main Results}
\label{sec:results}

\noindent\textbf{Selection is effective where the verifier is well matched.}
On CogVideoX-2B, progressive selection reaches 0.515, matching Best-of-$N$ while
scoring highest on S2 (0.91), the metric most directly measuring multi-frame
physical consistency, and improving over Random Selection (0.490).
The gap between Random selection and Base model shows that much of
the headline improvement comes from multi-trajectory sampling itself.
On Wan~2.1-14B selection gives the largest absolute gain
($0.569 \rightarrow 0.612$, $+7.6\%$), consistent across all four categories and
most pronounced on S2 ($+0.18$).
On CogVideoX-5B, by contrast, selection is essentially flat ($+0.002$): gains in
material properties ($+0.100$) are canceled by losses in optics and thermal.
GPT-4o pairwise judgments are nonetheless favorable (60.2\% of decided
comparisons against the base model, 65/43 with 45 ties over 153 valid pairs),
suggesting perceptual improvements the aggregate score does not capture.

\begin{figure*}[!t]
\centering
\setlength{\tabcolsep}{1pt}
\small
\newcommand{\qprompt}[1]{\parbox[t]{0.31\linewidth}{\centering\scriptsize\textit{#1}}}
\begin{tabular}{@{}r@{\hspace{3pt}} c @{\hspace{5pt}} c @{\hspace{5pt}} c@{}}
 & \textbf{Mechanics} & \textbf{Optics} & \textbf{Thermal} \\[2pt]
\rotatebox[origin=c]{90}{\small\textsf{Baseline}} &
\includegraphics[width=0.31\linewidth]{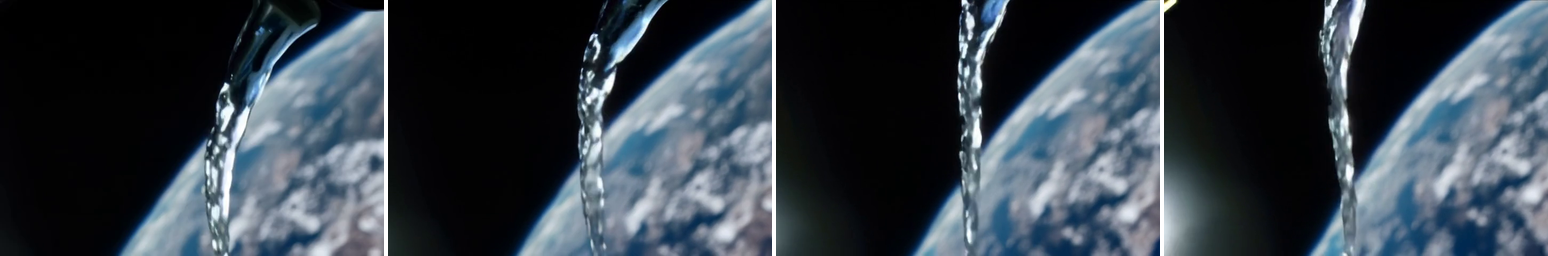} &
\includegraphics[width=0.31\linewidth]{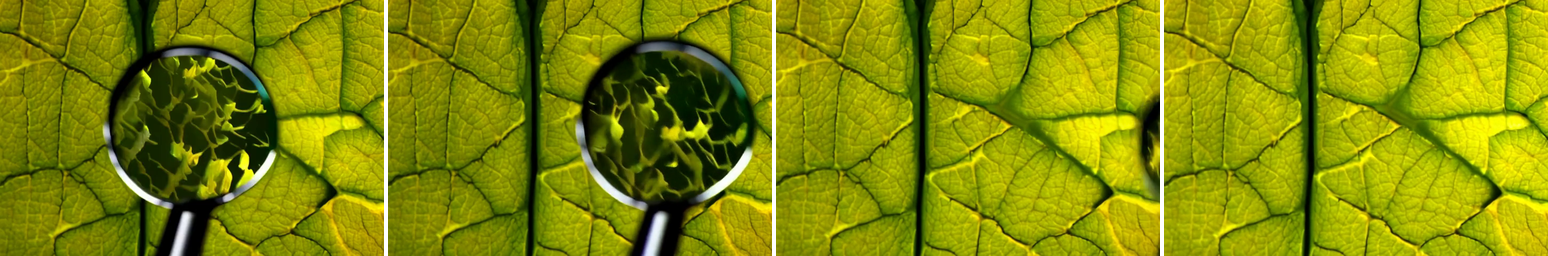} &
\includegraphics[width=0.31\linewidth]{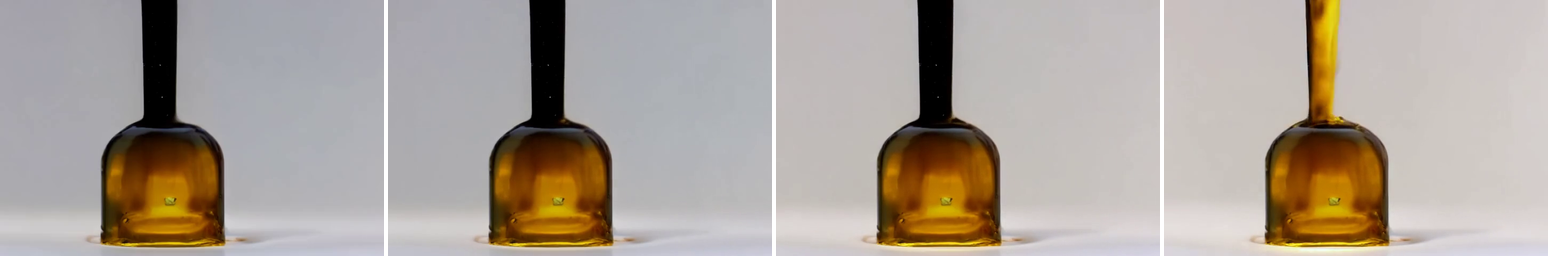} \\[2pt]
\rotatebox[origin=c]{90}{\small\textsf{Ours}} &
\includegraphics[width=0.31\linewidth]{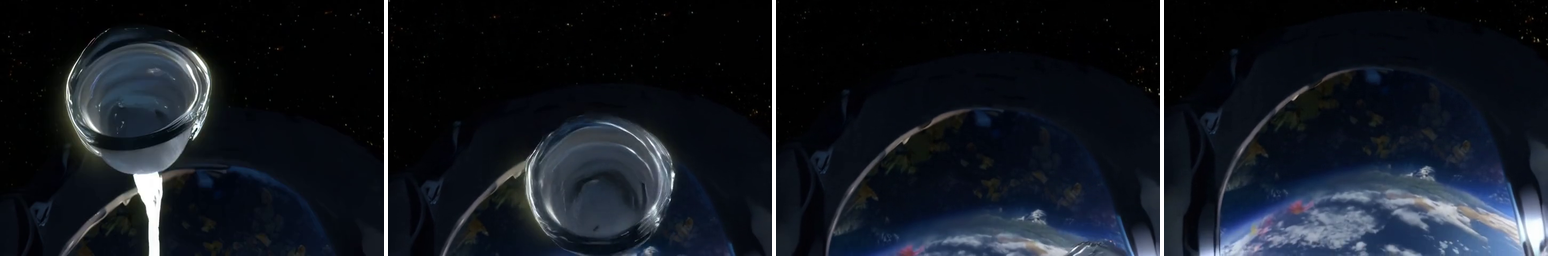} &
\includegraphics[width=0.31\linewidth]{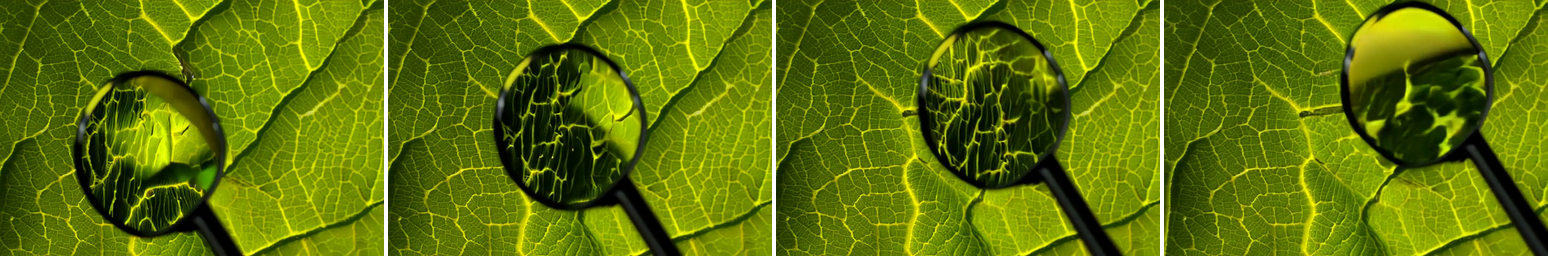} &
\includegraphics[width=0.31\linewidth]{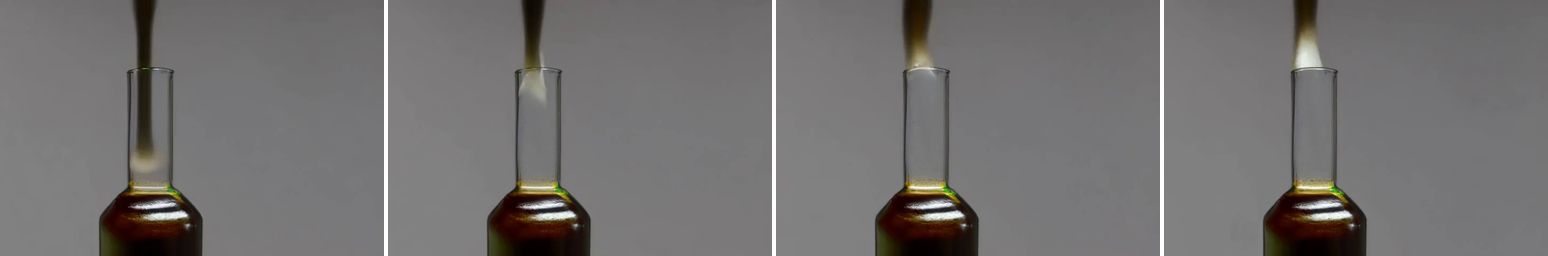} \\[2pt]
&
\qprompt{``A cup of oil is slowly poured out in the space station, releasing the liquid into the surrounding area''} &
\qprompt{``A magnifying glass is gradually moving closer to a leaf, revealing the intricate details and textures of the veins and surface patterns as it approaches.''} &
\qprompt{``A timelapse captures the transformation of arsenic trioxide as it is exposed to gradually increasing temperature at room temperature''} \\
\end{tabular}
\caption{\textbf{Qualitative comparison} on PhyGenBench prompts
(4~uniformly sampled frames per video).
\textbf{Mechanics}: baseline pours oil downward under terrestrial
gravity; ours forms a floating liquid mass consistent with
microgravity.
\textbf{Optics}: the texture magnified through the baseline's
lens appears incoherent with the surrounding leaf venation;
ours produces a magnified view whose structure is consistent
with the underlying leaf surface.
\textbf{Thermal and Physical Properties}: baseline shows a liquid stream (incorrect);
ours shows rising vapor consistent with sublimation.}
\label{fig:qualitative}
\end{figure*}

\noindent\textbf{Reward gradient helps most where selection cannot.}
The pattern inverts for gradient guidance.
On CogVideoX-5B it lifts the score from 0.363 to 0.496 ($+0.133$), with gains in
every category and the largest jump on S2 ($+0.37$).
That suggests when all $N$ initial samples are
comparably poor and re-ranking cannot help, guidance actively moves each
trajectory.
On Wan, adding guidance on top of selection brings no benefit
($0.612 \rightarrow 0.606$), and the GPT-4o judge ties guidance against the
base model on 157 of 160 prompts, i.e.\ no measurable effect on an already
strong backbone.

\noindent\textbf{DPO is not uniformly better.}
DPO reaches 0.475 on CogVideoX-5B, below reward-gradient guidance (0.496) despite
updating model weights.
On Wan it \emph{degrades} the base model
($0.569 \rightarrow 0.558$), driven almost entirely by a drop in naturalness
(S3: $1.79 \rightarrow 1.53$), a signal of reward over-optimization that
inference-time mechanisms avoid, since they never move the model off its own
distribution.

\noindent\textbf{Efficiency.}
Best-of-$N$ runs all four trajectories to completion, whereas selection terminates two at $t{=}600$ and one more at $t{=}400$.
On CogVideoX-2B this is a \textbf{37\%} wall-clock reduction (490s \vs\ 778s) at
matched output quality.\footnote{Single NVIDIA A5000; wall-clock includes prompt
encoding, denoising, and decoding.}
Checkpoint scoring adds one conditional forward per active trajectory
(${\approx}5\%$ of generation time, included in these timings); the verifier
head itself is negligible.
Reward-gradient guidance trades part of this saving for quality: its backward pass
reaches only the first $\ell_{\mathrm{best}}{=}10$ of $L$ blocks and runs at a few
guidance steps only, so the combined method remains cheaper than running all $N$
trajectories to completion.

\noindent\textbf{Where the gains come from.}
Score-distribution analysis (supplementary) shows kept and dropped trajectories
separate visibly on CogVideoX-5B but overlap substantially on Wan, indicating that
on Wan the selection gain is driven more by multi-trajectory diversity than by
physics-informed ranking.
That ranking contributes independently is established by the Random Selection
control, where random selection drops the score from 0.515 to 0.490, so the physics signal accounts for that margin.
The Wan case is instead limited by verifier strength, as the Wan
verifier is trained on only 591 annotated Wan videos (against 1{,}736 for CogVideoX-5B), and keeps a held-out AUC
of 0.657. However, at that level it is competitive with
training-free LikePhys scorer (\cref{tab:likephys}).


\begin{table}[t]
\centering
\resizebox{0.475\textwidth}{!}{%
\begin{tabular}{@{}lccc@{}}
\toprule
Method & Physics-IQ & Spatial IoU & MSE~$\downarrow$ \\
\midrule
Baseline            & 0.242 & 0.292 & 0.0064 \\
$+$ Reward Gradient & \textbf{0.262} & \textbf{0.317} & \textbf{0.0063} \\
\bottomrule
\end{tabular}
}
\caption{\textbf{Physics-IQ results} on CogVideoX-5B-I2V (42 of the 66
scenarios, all three camera perspectives; protocol in the supplementary).
The gain concentrates in spatial IoU ($0.292 \rightarrow 0.317$), which
scores the regions where the scene actually moves.}
\label{tab:physicsiq}
\end{table}

\noindent\textbf{Zero-shot transfer to I2V.}
Physics-IQ requires image-to-video generation, for which we use
CogVideoX-5B-I2V.
Since the I2V variant shares the DiT architecture of its T2V counterpart, we
apply the T2V-trained verifier \emph{zero-shot}: its AUC on I2V features
(0.662) matches the T2V value (0.660), and retraining natively on I2V features
brings no gain (0.643), so all Physics-IQ runs reuse the T2V verifier
unchanged.
With this transferred verifier, selection with reward gradient improves the
Physics-IQ score from 0.242 to 0.262 (Table~\ref{tab:physicsiq}), indicating
that the verifier signal survives the change of conditioning modality.


\begin{table}[t]
\centering
\resizebox{0.475\textwidth}{!}{%
\begin{tabular}{@{}lccc@{}}
\toprule
Verifier & CogX-2B & CogX-5B & Wan~2.1-14B \\
\midrule
LikePhys (training-free) & 0.554 & 0.488 & 0.627 \\
Ours (trained)           & 0.684 & 0.660 & 0.657 \\
\bottomrule
\end{tabular}
}
\caption{\textbf{Verifier AUC} on the held-out split of each verifier's
training annotations.}
\label{tab:likephys}
\vspace{-1em}
\end{table}

\section{Discussion and Conclusion}
\label{sec:discussion}

We have shown that frozen DiT intermediate features carry a
measurable signal predictive of physical plausibility.
This signal persists in within-source analyses and is not fully
explained by the coarse perceptual-quality proxy considered in
our controls.
A lightweight physics verifier trained on these features enables
progressive trajectory selection that matches Best-of-4 on
CogVideoX-2B while reducing wall-clock inference time by 37\%.
Reward-gradient guidance further improves physical consistency
on CogVideoX-5B.
Taken together, the probing analysis in the \secref{sec:probing} section
and the downstream results in the \secref{sec:experiments} section suggest
that video diffusion transformers acquire implicit physical
knowledge as a byproduct of learning to denoise, even without
explicit physics supervision.


Several limitations remain.
First, the probing signal is moderate, which limits the achievable
selection accuracy when candidate trajectories differ only subtly
in physical plausibility.
%
Second, the current framework requires training a separate
verifier for each target diffusion backbone, limiting plug-and-play
transfer across architectures.
Third, our verifier is trained on a limited set of labeled videos
and may not generalize to physical phenomena that are
underrepresented in the current dataset.
Finally, our evaluation follows PhyGenEval and relies on GPT-4o~\cite{openai2024gpt4o},
inheriting the noise and potential bias of VLM-based judging.


These limitations suggest directions for future work.
Scaling training data with synthetic or simulator-generated
scenarios could strengthen the signal and broaden physical coverage.
Extending the analysis to larger DiTs and additional backbones
would clarify how such representations scale with model capacity.
Finally, combining selection with lightweight test-time steering~\cite{liu2024physdreamer,yuan2025newtongen,gillman2025forceprompting}
may further close the gap to physics-aware fine-tuning while
preserving the modularity and efficiency of our approach.



\bibliography{main}

@String(CVPR  = {IEEE Conf. Comput. Vis. Pattern Recog.})

@String(ICCV  = {Int. Conf. Comput. Vis.})

@String(ECCV  = {Eur. Conf. Comput. Vis.})

@String(NeurIPS = {Adv. Neural Inform. Process. Syst.})

@String(ICML  = {Int. Conf. Mach. Learn.})

@String(ICLR  = {Int. Conf. Learn. Represent.})

@String(BMVC  = {Brit. Mach. Vis. Conf.})

@String(CVPR  = {CVPR})

@String(ICCV  = {ICCV})

@String(ECCV  = {ECCV})

@String(NeurIPS = {NeurIPS})

@String(ICML  = {ICML})

@String(ICLR  = {ICLR})

@String(BMVC  =	{BMVC})

@inproceedings{yang2025cogvideox,
  title     = {{CogVideoX}: Text-to-Video Diffusion Models with an Expert Transformer},
  author    = {Yang, Zhuoyi and Teng, Jiayan and Zheng, Wendi and Ding, Ming and Huang, Shiyu and Xu, Jiazheng and Yang, Yuanming and Hong, Wenyi and Zhang, Xiaohan and Feng, Guanyu and Yin, Da and Zhang, Yuxuan and Wang, Weihan and Cheng, Yean and Xu, Bin and Gu, Xiaotao and Dong, Yuxiao and Tang, Jie},
  booktitle = {International Conference on Learning Representations (ICLR)},
  year      = {2025}
}

@article{opensora2024,
  title   = {Open-{S}ora: Democratizing Efficient Video Production for All},
  author  = {Zheng, Zangwei and Peng, Xiangyu and Yang, Tianji and Shen, Chenhui and Li, Shenggui and Liu, Hongxin and Zhou, Yukun and Li, Tianyi and You, Yang},
  journal = {arXiv preprint arXiv:2412.20404},
  year    = {2024}
}

@article{wanteam2025wan,
  title   = {Wan: Open and Advanced Large-Scale Video Generative Models},
  author  = {{Team Wan} and Wang, Ang and Ai, Baole and Wen, Bin and Mao, Chaojie and Xie, Chen-Wei and Chen, Di and Yu, Feiwu and Zhao, Haiming and Yang, Jianxiao and Zeng, Jianyuan and Wang, Jiayu and Zhang, Jingfeng and Zhou, Jingren and Wang, Jinkai and Chen, Jixuan and Zhu, Kai and Zhao, Kang and Yan, Keyu and Huang, Lianghua and Feng, Mengyang and Zhang, Ningyi and Li, Pandeng and Wu, Pingyu and Chu, Ruihang and Feng, Ruili and Zhang, Shiwei and Sun, Siyang and Fang, Tao and Wang, Tianxing and Gui, Tianyi and Weng, Tingyu and Shen, Tong and Lin, Wei and Wang, Wei and Wang, Wei and Zhou, Wenmeng and Wang, Wente and Shen, Wenting and Yu, Wenyuan and Shi, Xianzhong and Huang, Xiaoming and Xu, Xin and Kou, Yan and Lv, Yangyu and Li, Yifei and Liu, Yijing and Wang, Yiming and Zhang, Yingya and Huang, Yitong and Li, Yong and Wu, You and Liu, Yu and Pan, Yulin and Zheng, Yun and Hong, Yuntao and Shi, Yupeng and Feng, Yutong and Jiang, Zeyinzi and Han, Zhen and Wu, Zhi-Fan and Liu, Ziyu},
  journal = {arXiv preprint arXiv:2503.20314},
  year    = {2025}
}

@article{kong2024hunyuanvideo,
  title   = {{HunyuanVideo}: A Systematic Framework For Large Video Generative Models},
  author  = {Kong, Weijie and Tian, Qi and Zhang, Zijian and Min, Rox and Dai, Zuozhuo and Zhou, Jin and Xiong, Jiangfeng and Li, Xin and Wu, Bo and Zhang, Jianwei and Wu, Kathrina and Lin, Qin and Yuan, Junkun and Long, Yanxin and Wang, Aladdin and Wang, Andong and Li, Changlin and Huang, Duojun and Yang, Fang and Tan, Hao and Wang, Hongmei and Song, Jacob and Bai, Jiawang and Wu, Jianbing and Xue, Jinbao and Wang, Joey and Wang, Kai and Liu, Mengyang and Li, Pengyu and Li, Shuai and Wang, Weiyan and Yu, Wenqing and Deng, Xinchi and Li, Yang and Chen, Yi and Cui, Yutao and Peng, Yuanbo and Yu, Zhentao and He, Zhiyu and Xu, Zhiyong and Zhou, Zixiang and Xu, Zunnan and Tao, Yangyu and Lu, Qinglin and Liu, Songtao and Zhou, Dax and Wang, Hongfa and Yang, Yong and Wang, Di and Liu, Yuhong and Jiang, Jie and Zhong, Caesar},
  journal = {arXiv preprint arXiv:2412.03603},
  year    = {2024}
}

@misc{li2025pika,
  title  = {Pika},
  author = {{Pika Labs}},
  year   = {2023},
  note   = {\url{https://pika.art}}
}

@inproceedings{wang2025wisa,
  title={WISA: World simulator assistant for physics-aware text-to-video generation},
  author={Wang, Jing and Ma, Ao and Cao, Ke and Zheng, Jun and Feng, Jiasong and Zhang, Zhanjie and Pang, Wanyuan and Liang, Xiaodan},
  booktitle={The Thirty-ninth Annual Conference on Neural Information Processing Systems},
  year    = {2025}
}

@article{zhang2025thinkbefore,
  title   = {Think Before You Diffuse: Infusing Physical Rules into Video Diffusion},
  author  = {Zhang, Ke and Xiao, Cihan and Xu, Jiacong and Mei, Yiqun and Patel, Vishal M.},
  journal = {arXiv preprint arXiv:2505.21653},
  year    = {2025}
}

@inproceedings{zhang2025videorepa,
  title     = {VideoREPA: Learning Physics for Video Generation through Relational Alignment with Foundation Models},
  author    = {Zhang, Xiangdong and Liao, Jiaqi and Zhang, Shaofeng and Meng, Fanqing and Wan, Xiangpeng and Yan, Junchi and Cheng, Yu},
  booktitle = {Advances in Neural Information Processing Systems},
  year      = {2025}
}

@inproceedings{wu2024boosting,
  title     = {Boosting Text-to-Video Generative Model with MLLMs Feedback},
  author    = {Wu, Xun and Huang, Shaohan and Wang, Guolong and Xiong, Jing and Wei, Furu},
  booktitle = {Advances in Neural Information Processing Systems},
  year      = {2024}
}

@inproceedings{liu2025videoreward,
  title     = {Improving Video Generation with Human Feedback},
  author    = {Liu, Jie and Liu, Gongye and Liang, Jiajun and Yuan, Ziyang and Liu, Xiaokun and Zheng, Mingwu and Wu, Xiele and Wang, Qiulin and Xia, Menghan and Wang, Xintao and Liu, Xiaohong and Yang, Fei and Wan, Pengfei and Zhang, Di and Gai, Kun and Yang, Yujiu and Ouyang, Wanli},
  booktitle = {Advances in Neural Information Processing Systems},
  year      = {2025},
  note      = {arXiv:2501.13918}
}

@inproceedings{oshima2025dlbs,
  title     = {Inference-Time Text-to-Video Alignment with Diffusion Latent Beam Search},
  author    = {Oshima, Yuta and Suzuki, Masahiro and Matsuo, Yutaka and Furuta, Hiroki},
  booktitle = {Advances in Neural Information Processing Systems},
  year      = {2025},
  note      = {arXiv:2501.19252}
}

@inproceedings{yuan2025newtongen,
  title     = {NewtonGen: Physics-Consistent and Controllable Text-to-Video Generation via Neural Newtonian Dynamics},
  author    = {Yuan, Yu and Wang, Xijun and Wickremasinghe, Tharindu and Nadir, Zeeshan and Ma, Bole and Chan, Stanley H.},
  booktitle = {International Conference on Learning Representations (ICLR)},
  year      = {2026},
  note      = {arXiv:2509.21309}
}

@inproceedings{liu2024physdreamer,
  title     = {{PhysDreamer}: Physics-Based Interaction with 3{D} Objects via Video Generation},
  author    = {Zhang, Tianyuan and Yu, Hong-Xing and Wu, Rundi and Feng, Brandon Y. and Zheng, Changxi and Snavely, Noah and Wu, Jiajun and Freeman, William T.},
  booktitle = {European Conference on Computer Vision (ECCV)},
  year      = {2024},
  note      = {arXiv:2404.13026}
}

@inproceedings{zhang2025tora,
  author    = {Zhang, Zhenghao and Liao, Junchao and Li, Menghao and Dai, Zuozhuo and Qiu, Bingxue and Zhu, Siyu and Qin, Long and Wang, Weizhi},
  title     = {Tora: Trajectory-oriented Diffusion Transformer for Video Generation},
  booktitle = {Proceedings of the IEEE/CVF Conference on Computer Vision and Pattern Recognition (CVPR)},
  year      = {2025},
}

@inproceedings{akkerman2025interdyn,
  author    = {Akkerman, Rick and Feng, Haiwen and Black, Michael J. and Tzionas, Dimitrios and Abrevaya, Victoria Fern{\'a}ndez},
  title     = {InterDyn: Controllable Interactive Dynamics with Video Diffusion Models},
  booktitle = {Proceedings of the IEEE/CVF Conference on Computer Vision and Pattern Recognition (CVPR)},
  year      = {2025},
}

@inproceedings{geng2025motionprompting,
  author    = {Geng, Daniel and Herrmann, Charles and Hur, Junhwa and Cole, Forrester and Zhang, Serena and Pfaff, Tobias and Lopez-Guevara, Tatiana and Doersch, Carl and Aytar, Yusuf and Rubinstein, Michael and Sun, Chen and Wang, Oliver and Owens, Andrew and Sun, Deqing},
  title     = {Motion Prompting: Controlling Video Generation with Motion Trajectories},
  booktitle = {Proceedings of the IEEE/CVF Conference on Computer Vision and Pattern Recognition (CVPR)},
  year      = {2025},
}

@inproceedings{gillman2025forceprompting,
  title     = {Force Prompting: Video Generation Models Can Learn and Generalize Physics-based Control Signals},
  author    = {Gillman, Nate and Herrmann, Charles and Freeman, Michael and Aggarwal, Daksh and Luo, Evan and Sun, Deqing and Sun, Chen},
  booktitle = {Advances in Neural Information Processing Systems},
  year      = {2025},
  note      = {arXiv:2505.19386}
}

@inproceedings{gillman2026goalforce,
  title     = {Goal Force: Teaching Video Models To Accomplish Physics-Conditioned Goals},
  author    = {Gillman, Nate and Zhou, Yinghua and Tang, Zitian and Luo, Evan and Chakravarthy, Arjan and Aggarwal, Daksh and Freeman, Michael and Herrmann, Charles and Sun, Chen},
  booktitle = {Proceedings of the IEEE/CVF Conference on Computer Vision and Pattern Recognition (CVPR)},
  year      = {2026},
  note      = {arXiv:2601.05848}
}

@inproceedings{wang2025physctrl,
  title     = {{PhysCtrl}: Generative Physics for Controllable and Physics-Grounded Video Generation},
  author    = {Wang, Chen and Chen, Chuhao and Huang, Yiming and Dou, Zhiyang and Liu, Yuan and Gu, Jiatao and Liu, Lingjie},
  booktitle = {Advances in Neural Information Processing Systems},
  year      = {2025},
  note      = {arXiv:2509.20358}
}

@article{romero2025kinemask,
  title   = {Learning to Generate Rigid Body Interactions with Video Diffusion Models},
  author  = {Romero, David and Bermudez, Ariana and Iablochnikov, Viacheslav and Li, Hao and Pizzati, Fabio and Laptev, Ivan},
  journal = {arXiv preprint arXiv:2510.02284},
  year    = {2025}
}

@inproceedings{he2025evosearch,
  title     = {Scaling Image and Video Generation via Test-Time Evolutionary Search},
  author    = {He, Haoran and Liang, Jiajun and Wang, Xintao and Wan, Pengfei and Zhang, Di and Gai, Kun and Pan, Ling},
  booktitle = {Advances in Neural Information Processing Systems},
  year      = {2025},
  note      = {arXiv:2505.17618}
}

@inproceedings{liu2024physgen,
  title     = {{PhysGen}: Rigid-Body Physics-Grounded Image-to-Video Generation},
  author    = {Liu, Shaowei and Ren, Zhongzheng and Gupta, Saurabh and Wang, Shenlong},
  booktitle = {European Conference on Computer Vision (ECCV)},
  year      = {2024},
  note      = {arXiv:2409.18964}
}

@inproceedings{le2025vlipp,
  title     = {{VLIPP}: Towards Physically Plausible Video Generation with Vision and Language Informed Physical Prior},
  author    = {Yang, Xindi and Li, Baolu and Zhang, Yiming and Yin, Zhenfei and Bai, Lei and Ma, Liqian and Wang, Zhiyong and Cai, Jianfei and Wong, Tien-Tsin and Lu, Huchuan and Jia, Xu},
  booktitle = {International Conference on Computer Vision (ICCV)},
  year      = {2025}
}

@inproceedings{li2025pisaexperiments,
  title     = {Pisa Experiments: Exploring Physics Post-Training for Video Diffusion Models by Watching Stuff Drop},
  author    = {Li, Chenyu and Michel, Oscar and Pan, Xichen and Liu, Sainan and Roberts, Mike and Xie, Saining},
  booktitle = {International Conference on Machine Learning (ICML)},
  year      = {2025},
  note      = {arXiv:2503.09595}
}

@inproceedings{cai2025phygdpo,
  title     = {{PhyGDPO}: Physics-Aware Groupwise Direct Preference Optimization for Physically Consistent Text-to-Video Generation},
  author    = {Cai, Yuanhao and Li, Kunpeng and Jia, Menglin and Wang, Jialiang and Sun, Junzhe and Liang, Feng and Chen, Weifeng and Juefei-Xu, Felix and Wang, Chu and Thabet, Ali and Dai, Xiaoliang and Ju, Xuan and Yuille, Alan and Hou, Ji},
  booktitle = {European Conference on Computer Vision (ECCV)},
  year      = {2026}
}

@article{prabhudesai2024aligningvideo,
  title   = {Video Diffusion Alignment via Reward Gradients},
  author  = {Prabhudesai, Mihir and Mendonca, Russell and Qin, Zheyang and Fragkiadaki, Katerina and Pathak, Deepak},
  journal = {arXiv preprint arXiv:2407.08737},
  year    = {2024}
}

@inproceedings{yuan2024instructvideo,
  title     = {{InstructVideo}: Instructing Video Diffusion Models with Human Feedback},
  author    = {Yuan, Hangjie and Zhang, Shiwei and Wang, Xiang and Wei, Yujie and Feng, Tao and Pan, Yining and Zhang, Yingya and Liu, Ziwei and Albanie, Samuel and Ni, Dong},
  booktitle = {IEEE/CVF Conference on Computer Vision and Pattern Recognition (CVPR)},
  year      = {2024},
  note      = {arXiv:2312.12490}
}

@inproceedings{meng2025phygenbench,
  title     = {Towards World Simulator: Crafting Physical Commonsense-Based Benchmark for Video Generation},
  author    = {Meng, Fanqing and Liao, Jiaqi and Tan, Xinyu and Shao, Wenqi and Lu, Quanfeng and Zhang, Kaipeng and Cheng, Yu and Li, Dianqi and Qiao, Yu and Luo, Ping},
  booktitle = {International Conference on Machine Learning (ICML)},
  year      = {2025}
}

@misc{bansal2024videophy,
  title         = {{VideoPhy}: Evaluating Physical Commonsense for Video Generation},
  author        = {Bansal, Hritik and Lin, Zongyu and Xie, Tianyi and Zong, Zeshun and Yarom, Michal and Bitton, Yonatan and Jiang, Chenfanfu and Sun, Yizhou and Chang, Kai-Wei and Grover, Aditya},
  year          = {2024},
  eprint        = {2406.03520},
  archivePrefix = {arXiv},
}

@misc{bansal2025videophy2,
  title         = {{VideoPhy-2}: A Challenging Action-Centric Physical Commonsense Evaluation in Video Generation},
  author        = {Bansal, Hritik and Peng, Clark and Bitton, Yonatan and Goldenberg, Roman and Grover, Aditya and Chang, Kai-Wei},
  year          = {2025},
  eprint        = {2503.06800},
  archivePrefix = {arXiv},
  primaryClass  = {cs.CV}
}

@inproceedings{wallace2024diffusiondpo,
  title={Diffusion Model Alignment Using Direct Preference Optimization},
  author={Wallace, Bram and Dang, Meihua and Rafailov, Rafael and Zhou, Linqi
          and Lou, Aaron and Purushwalkam, Senthil and Ermon, Stefano and
          Xiong, Caiming and Joty, Shafiq and Naik, Nikhil},
  booktitle={CVPR},
  year={2024}
}

@inproceedings{dhariwal2021classifier,
  title     = {Diffusion Models Beat {GAN}s on Image Synthesis},
  author    = {Dhariwal, Prafulla and Nichol, Alexander},
  booktitle = {Advances in Neural Information Processing Systems (NeurIPS)},
  year      = {2021}
}

@misc{bai2025qwen25vl,
  title         = {{Qwen2.5-VL} Technical Report},
  author        = {Bai, Shuai and Chen, Keqin and Liu, Xuejing and Wang, Jialin and Ge, Wenbin and Song, Sibo and Dang, Kai and Wang, Peng and Wang, Shijie and Tang, Jun and Zhong, Humen and Zhu, Yuanzhi and Yang, Mingkun and Li, Zhaohai and Wan, Jianqiang and Wang, Pengfei and Ding, Wei and Fu, Zheren and Xu, Yiheng and Ye, Jiabo and Zhang, Xi and Xie, Tianbao and Cheng, Zesen and Zhang, Hang and Yang, Zhibo and Xu, Haiyang and Lin, Junyang},
  year          = {2025},
  eprint        = {2502.13923},
  archivePrefix = {arXiv},
  primaryClass  = {cs.CV}
}

@misc{openai2024gpt4o,
  title        = {{GPT-4o} System Card},
  author       = {{OpenAI}},
  year         = {2024},
  howpublished = {https://openai.com/index/gpt-4o-system-card}
}

@inproceedings{lin2024evaluating,
  title={Evaluating text-to-visual generation with image-to-text generation},
  author={Lin, Zhiqiu and Pathak, Deepak and Li, Baiqi and Li, Jiayao and Xia, Xide and Neubig, Graham and Zhang, Pengchuan and Ramanan, Deva},
  booktitle={European Conference on Computer Vision},
  year={2024},
}

@article{svd2023stability,
  title={Stable Video Diffusion: Scaling Latent Video Diffusion Models to Large Datasets},
  author={Blattmann, Andreas and Dockhorn, Tim and Kulal, Sumith and Mendelevitch, Daniel and Kilian, Maciej and Lorenz, Dominik and Levi, Yam and English, Zion and Voleti, Vikram and Letts, Adam and Jampani, Varun and Rombach, Robin},
  journal={arXiv preprint arXiv:2311.15127},
  year={2023}
}

@misc{runwaygen2,
  title = {Runway Gen-2: {Text} to {Video} {Generation}},
  author = {RunwayML},
  year = {2023},
  howpublished = {\url{https://runwayml.com/research/gen-2}},
  note = {Accessed: 2026-03-04}
}

@misc{cerspense2023zeroscope,
  author = {Cerspense},
  title = {zeroscope\_v2: A watermark-free Modelscope-based video model},
  year = {2023},
  publisher = {Hugging Face},
  journal = {Hugging Face Repository},
  howpublished = {\url{https://huggingface.co/cerspense/zeroscope_v2_576w}},
}

@inproceedings{chen2024videocrafter2,
  title={Videocrafter2: Overcoming data limitations for high-quality video diffusion models},
  author={Chen, Haoxin and Zhang, Yong and Cun, Xiaodong and Xia, Menghan and Wang, Xintao and Weng, Chao and Shan, Ying},
  booktitle={Proceedings of the IEEE/CVF conference on computer vision and pattern recognition},
  year={2024}
}

@inproceedings{rombach2022high,
  title={High-resolution image synthesis with latent diffusion models},
  author={Rombach, Robin and Blattmann, Andreas and Lorenz, Dominik and Esser, Patrick and Ommer, Bj{\"o}rn},
  booktitle={Proceedings of the IEEE/CVF conference on computer vision and pattern recognition},
  year={2022}
}

@article{wang2025lavie,
  title     = {{LaVie}: High-Quality Video Generation with Cascaded Latent Diffusion Models},
  author    = {Wang, Yaohui and Chen, Xinyuan and Ma, Xin and Zhou, Shangchen and Huang, Ziqi and Wang, Yi and Yang, Ceyuan and He, Yinan and Yu, Jiashuo and Yang, Peiqing and Guo, Yuwei and Wu, Tianxing and Si, Chenyang and Jiang, Yuming and Chen, Cunjian and Loy, Chen Change and Dai, Bo and Lin, Dahua and Qiao, Yu and Liu, Ziwei},
  journal   = {International Journal of Computer Vision},
  volume    = {133},
  pages     = {3059--3078},
  year      = {2025},
  publisher = {Springer}
}

@inproceedings{wang2021causal,
  title={Causal attention for unbiased visual recognition},
  author={Wang, Tan and Zhou, Chang and Sun, Qianru and Zhang, Hanwang},
  booktitle={Proceedings of the IEEE/CVF international conference on computer vision},
  year={2021}
}

@misc{he2024videoscore,
  title         = {{VideoScore}: Building Automatic Metrics to Simulate Fine-grained Human Feedback for Video Generation},
  author        = {He, Xuan and Jiang, Dongfu and Zhang, Ge and Ku, Max and Soni, Achint and Siu, Sherman and Chen, Haonan and Chandra, Abhranil and Jiang, Ziyan and Arulraj, Aaran and Wang, Kai and Do, Quy Duc and Ni, Yuansheng and Lyu, Bohan and Narsupalli, Yaswanth and Fan, Rongqi and Lyu, Zhiheng and Lin, Yuchen and Chen, Wenhu},
  year          = {2024},
  eprint        = {2406.15252},
  archivePrefix = {arXiv},
}

@inproceedings{kang2025howfar,
  title     = {How Far Is Video Generation from World Model: {A} Physical Law Perspective},
  author    = {Kang, Bingyi and Yue, Yang and Lu, Rui and Lin, Zhijie and Zhao, Yang and Wang, Kaixin and Huang, Gao and Feng, Jiashi},
  booktitle = {International Conference on Machine Learning (ICML)},
  year      = {2025}
}

@misc{polyak2024moviegen,
  title         = {Movie Gen: A Cast of Media Foundation Models},
  author        = {Polyak, Adam and Zohar, Amit and Brown, Andrew and Tjandra, Andros and Sinha, Animesh and Lee, Ann and Vyas, Apoorv and Shi, Bowen and Ma, Chih-Yao and Chuang, Ching-Yao and Yan, David and Choudhary, Dhruv and Wang, Dingkang and Sethi, Geet and Pang, Guan and Ma, Haoyu and Misra, Ishan and Hou, Ji and Wang, Jialiang and Jagadeesh, Kiran and Li, Kunpeng and Zhang, Luxin and Singh, Mannat and Williamson, Mary and Le, Matt and Yu, Matthew and Singh, Mitesh Kumar and Zhang, Peizhao and Vajda, Peter and Duval, Quentin and Girdhar, Rohit and Sumbaly, Roshan and Rambhatla, Sai Saketh and Tsai, Sam and Azadi, Samaneh and Datta, Samyak and Chen, Sanyuan and Bell, Sean and Ramaswamy, Sharadh and Sheynin, Shelly and Bhattacharya, Siddharth and Motwani, Simran and Xu, Tao and Li, Tianhe and Hou, Tingbo and Hsu, Wei-Ning and Yin, Xi and Dai, Xiaoliang and Taigman, Yaniv and Luo, Yaqiao and Liu, Yen-Cheng and Wu, Yi-Chiao and Zhao, Yue and Kirstain, Yuval and He, Zecheng and He, Zijian and Pumarola, Albert and Thabet, Ali and Sanakoyeu, Artsiom and Mallya, Arun and Guo, Baishan and Araya, Boris and Kerr, Breena and Wood, Carleigh and Liu, Ce and Peng, Cen and Vengertsev, Dimitry and Schonfeld, Edgar and Blanchard, Elliot and Juefei-Xu, Felix and Nord, Fraylie and Liang, Jeff and Hoffman, John and Kohler, Jonas and Fire, Kaolin and Sivakumar, Karthik and Chen, Lawrence and Yu, Licheng and Gao, Luya and Georgopoulos, Markos and Moritz, Rashel and Sampson, Sara K. and Li, Shikai and Parmeggiani, Simone and Fine, Steve and Fowler, Tara and Petrovic, Vladan and Du, Yuming},
  year          = {2024},
  eprint        = {2410.13720},
  archivePrefix = {arXiv}
}

@misc{brooks2024sora,
  title        = {Video Generation Models as World Simulators},
  author       = {Brooks, Tim and Peebles, Bill and Holmes, Connor and DePue, Will and Guo, Yufei and Jing, Li and Schnurr, David and Taylor, Joe and Luhman, Troy and Luhman, Eric and Ng, Clarence and Wang, Ricky and Ramesh, Aditya},
  year         = {2024},
  howpublished = {\\url{https://openai.com/research/video-generation-models-as-world-simulators}}
}

@inproceedings{baraldi2025verifier,
  title     = {Verifier Matters: Enhancing Inference-Time Scaling for Video Diffusion Models},
  author    = {Baraldi, Lorenzo and Bucciarelli, Davide and Zeng, Zifan and Zhang, Chongzhe and Zhang, Qunli and Cornia, Marcella and Baraldi, Lorenzo and Liu, Feng and Hu, Zheng and Cucchiara, Rita},
  booktitle = {Proceedings of the British Machine Vision Conference (BMVC)},
  year      = {2025}
}

@inproceedings{raghu2021vision,
  title={Do Vision Transformers See Like Convolutional Neural Networks?},
  author={Raghu, Maithra and Unterthiner, Thomas and Kornblith, Simon and Zhang, Chiyuan and Dosovitskiy, Alexey},
  booktitle={NeurIPS},
  year={2021}
}

@inproceedings{caron2021emerging,
  title={Emerging Properties in Self-Supervised Vision Transformers},
  author={Caron, Mathilde and Touvron, Hugo and Misra, Ishan and J{\'e}gou, Herv{\'e} and Mairal, Julien and Bojanowski, Piotr and Joulin, Armand},
  booktitle={ICCV},
  year={2021}
}

@article{ho2022classifier,
  title={Classifier-free diffusion guidance},
  author={Ho, Jonathan and Salimans, Tim},
  journal={arXiv preprint arXiv:2207.12598},
  year={2022}
}

@article{esmati2026invisible,
  title={The Invisible Hand of Physics: When Video Diffusion Models Know More Than They Show},
  author={Esmati, Parsa and Nath, Somjit and Hofmann, Katja and Nowrouzezahrai, Derek and Kahou, Samira Ebrahimi and Mirmehdi, Majid},
  journal={arXiv preprint arXiv:2606.05328},
  year={2026}
}

@article{punzo2026layerwise,
  title={Do Video Foundation Models Understand Intuitive Physics? A Layerwise Probing Analysis},
  author={Punzo, Samuele and Caselli, Niccol{\`o} and Pantelidis, Ippokratis and Massafra, Francesco and Lo Sardo, Salvatore and Salehi, Mohammadreza},
  journal={arXiv preprint arXiv:2606.09646},
  year={2026}
}

@inproceedings{yuan2025likephys,
  title     = {LikePhys: Evaluating Intuitive Physics Understanding in Video Diffusion Models via Likelihood Preference},
  author    = {Yuan, Jianhao and Pizzati, Fabio and Pinto, Francesco and Kunze, Lars and Laptev, Ivan and Newman, Paul and Torr, Philip and De Martini, Daniele},
  booktitle = {International Conference on Learning Representations (ICLR)},
  year      = {2026},
  note      = {arXiv:2510.11512}
}

@article{yuan2026wmreward,
  title={Inference-time Physics Alignment of Video Generative Models with Latent World Models},
  author={Yuan, Jianhao and Zhang, Xiaofeng and Friedrich, Felix and Beltran-Velez, Nicolas and Hall, Melissa and Askari-Hemmat, Reyhane and Han, Xiaochuang and Ballas, Nicolas and Drozdzal, Michal and Romero-Soriano, Adriana},
  journal={arXiv preprint arXiv:2601.10553},
  year={2026}
}

@article{motamed2025physicsiq,
  title={Do Generative Video Models Understand Physical Principles?},
  author={Motamed, Saman and Culp, Laura and Swersky, Kevin and Jaini, Priyank and Geirhos, Robert},
  journal={arXiv preprint arXiv:2501.09038},
  year={2025}
}

@article{garrido2025intuitive,
  title={Intuitive Physics Understanding Emerges from Self-Supervised Pretraining on Natural Videos},
  author={Garrido, Quentin and Ballas, Nicolas and Assran, Mahmoud and Bardes, Adrien and Najman, Laurent and Rabbat, Michael and Dupoux, Emmanuel and LeCun, Yann},
  journal={arXiv preprint arXiv:2502.11831},
  year={2025}
}

@inproceedings{liu2025videot1,
  title     = {Video-T1: Test-Time Scaling for Video Generation},
  author    = {Liu, Fangfu and Wang, Hanyang and Cai, Yimo and Zhang, Kaiyan and Zhan, Xiaohang and Duan, Yueqi},
  booktitle = {International Conference on Computer Vision (ICCV)},
  year      = {2025},
  note      = {arXiv:2503.18942}
}

\clearpage
\appendix

\setcounter{figure}{0}
\setcounter{table}{0}
\setcounter{equation}{0}
\renewcommand{\thefigure}{S\arabic{figure}}
\renewcommand{\thetable}{S\arabic{table}}
\renewcommand{\theequation}{S\arabic{equation}}

\makeatletter
\setlength{\@fptop}{0pt}
\setlength{\@fpsep}{8pt}
\setlength{\@fpbot}{0pt plus 1fil}
\makeatother
\renewcommand{\floatpagefraction}{0.85}
\renewcommand{\topfraction}{0.9}
\renewcommand{\textfraction}{0.05}

\begin{center}
    {\Large\bf Supplementary Material}
\end{center}

\noindent
This supplementary document provides additional implementation details
(\cf~\secref{sec:supp_impl}), additional experimental analysis
(\cf~\secref{sec:supp_experiments}), qualitative examples
(\cf~\secref{sec:visualization}), and failure cases
(\cf~\secref{sec:supp_failure}) that complement the main paper.

\section{Additional Implementation Details}
\label{sec:supp_impl}

This section provides additional details on the physics verifier
architecture, feature extraction pipeline, and video generation setup,
complementing the method and experiment sections of the main paper. As
reported in the main paper, beyond the primary experiments on
CogVideoX-2B~\cite{yang2025cogvideox} we further evaluate our framework
on CogVideoX-5B~\cite{yang2025cogvideox} and
Wan~2.1-14B~\cite{wanteam2025wan}. The physics verifier architecture is
shared across all three settings, with only the input projection layer
adapted to the hidden dimension of each backbone.

\subsection{Physics Verifier Architecture}
\label{sec:supp_arch}

The physics verifier takes as input the spatially pooled per-frame DiT
features extracted from layer $\ell$ at denoising timestep $t$, denoted
by $\mathbf{f}^{(\ell)}_t \in \mathbb{R}^{F \times D}$, where $F=13$ is
the number of latent frames and $D$ is the hidden dimension of the
backbone. \Cref{tab:verifier_arch} presents the full layer-by-layer
architecture, while \cref{tab:backbone_specs} summarizes the
backbone-specific configurations.

\begin{table}[h]
\centering
\caption{Architecture of the proposed physics verifier. The input
projection layer is adapted to the hidden dimension $D$ of the
corresponding backbone, while all remaining components are shared
across backbones.}
\label{tab:verifier_arch}
\setlength{\tabcolsep}{3pt}
\resizebox{\linewidth}{!}{
\begin{tabular}{@{}lll@{}}
\toprule
\textbf{Component} & \textbf{Operation} & \textbf{Output Shape} \\
\midrule
Input & Spatially pooled per-frame DiT features & $[B, F, D]$ \\
Input Projection & $\mathrm{Linear}(D \rightarrow d)$ & $[B, F, d]$ \\
Positional Embedding & Add learnable $\mathbf{p}_{1:F} \in \mathbb{R}^{F \times d}$ & $[B, F, d]$ \\
Layer Normalization & Pre-normalization & $[B, F, d]$ \\
Causal Self-Attention & Multi-head ($h = 4$) with a causal mask & $[B, F, d]$ \\
Residual Connection & $\tilde{\mathbf{f}} + \mathrm{CausalAttn}(\mathrm{LN}(\tilde{\mathbf{f}}))$ & $[B, F, d]$ \\
Sequence Pooling & Select the final-frame representation & $[B, d]$ \\
Layer Normalization & Apply LayerNorm & $[B, d]$ \\
Classifier Head & $\mathrm{Lin}(d {\rightarrow} 128) {\rightarrow} \mathrm{GELU} {\rightarrow} \mathrm{Lin}(128 {\rightarrow} 1)$ & $[B, 1]$ \\
Output & Sigmoid physical plausibility score & $[B, 1]$ \\
\bottomrule
\end{tabular}
}
\end{table}

\begin{table}[h]
\centering
\caption{Backbone-specific configurations of the three DiT backbones
evaluated in this work. The physics verifier architecture ($d = 256$,
$h = 4$) is shared across all settings; only the input projection layer
and the feature extraction hook differ across backbones.}
\label{tab:backbone_specs}
\setlength{\tabcolsep}{3pt}
\resizebox{\linewidth}{!}{%
\small
\begin{tabular}{@{}lccc@{}}
\toprule
& \textbf{CogVideoX-2B} & \textbf{CogVideoX-5B} & \textbf{Wan~2.1-14B} \\
\midrule
Hidden dimension $D$ & 1920 & 3072 & 5120 \\
Number of transformer layers & 30 & 42 & 40 \\
Text tokens in sequence & Yes (226 tokens) & Yes (226 tokens) & No (cross-attention) \\
Feature extraction hook & \texttt{transformer\_blocks} & \texttt{transformer\_blocks} & \texttt{blocks} \\
Positional encoding & Learned & Rotary (RoPE) & --- \\
Sampling scheduler & DDIM & DDIM & UniPCMultistep \\
\midrule
Verifier parameters & $\sim 0.55$M & $\sim 0.85$M & $\sim 1.37$M \\
Verifier input shape & $[13, 1920]$ & $[13, 3072]$ & $[13, 5120]$ \\
\bottomrule
\end{tabular}%
}
\end{table}

\noindent\textbf{Design choices.}
The causal self-attention mask ensures that the representation at
frame~$i$ attends only to frames $j \leq i$, thereby enforcing temporal
causality. This design is motivated by the observation that physical
violations often emerge as progressively accumulated inconsistencies
over time (\eg, an object gradually deviating from a physically
plausible trajectory across successive frames). Using the final-frame
representation as the sequence summary allows the verifier to aggregate
information from all preceding frames while preserving causal ordering.
We use $d = 256$ and $h = 4$ attention heads in all experiments. The
parameter count is dominated by the input projection layer,
$\mathrm{Linear}(D \rightarrow d)$, and therefore scales with the
backbone hidden dimension $D$: from approximately $0.55$M parameters
for CogVideoX-2B ($D = 1920$) to approximately $1.37$M for Wan~2.1-14B
($D = 5120$). In every setting, the verifier adds at most
${\sim}0.03\%$ additional parameters relative to its backbone.

\noindent\textbf{Backbone-specific differences in feature extraction.}
The three backbones differ primarily in their text-conditioning
mechanisms. CogVideoX-2B and CogVideoX-5B prepend 226 T5 text tokens to
the video-token sequence, and we remove these text tokens during
feature extraction. By contrast, Wan~2.1-14B uses cross-attention for
text conditioning, so its hidden states contain only video tokens. In
addition, CogVideoX-5B requires explicit rotary positional embeddings
(\texttt{image\_rotary\_emb}) at each transformer forward pass;
otherwise, generation collapses to all-black outputs. Finally, because
the UniPCMultistep scheduler used by Wan~2.1-14B is stateful, we
maintain a separate deep-copied scheduler for each parallel trajectory
during multi-trajectory generation.

\subsection{Training Details}
\label{sec:supp_training}

\noindent\textbf{Training data.}
Following the matched-distribution principle, each verifier is trained only on
videos generated by its own backbone: 343 CogVideoX-2B generations with binary
physical-commonsense labels from VideoPhy~\cite{bansal2024videophy}, 1{,}736
CogVideoX-5B generations, and 591 Wan~2.1 generations with 5-point PC ratings
from VideoPhy-2~\cite{bansal2025videophy2}, binarized at
$\text{PC}{\ge}3$.
Each pool is split 85\%/15\% into train/validation (seed 42).
For every video we extract features with the corresponding frozen backbone at
$t \in \{200, 400, 600\}$, tripling the number of training samples; features
are stored as \texttt{fp16} tensors of shape $[13, D]$.

\vspace{0.5em}
\noindent\textbf{Training hyperparameters.}
\Cref{tab:training_hyper} summarizes the training configuration, which
is kept identical across all three backbones. To address class
imbalance (36.9\% positive for CogVideoX-2B, 66.4\% for CogVideoX-5B), we use
weighted binary cross-entropy with $\mathtt{pos\_weight} = n_{\mathrm{neg}} /
n_{\mathrm{pos}}$. Training completes in under 10 minutes on a single
GPU for each backbone.

\begin{table}[t!]
\centering
\caption{Training hyperparameters of the physics verifier, shared
across all backbones.}
\label{tab:training_hyper}
\setlength{\tabcolsep}{6pt}
\resizebox{\linewidth}{!}{%
\begin{tabular}{@{}ll@{}}
\toprule
\textbf{Hyperparameter} & \textbf{Value} \\
\midrule
Optimizer & AdamW \\
Learning rate & $1 \times 10^{-3}$ \\
Weight decay & 0.01 \\
Batch size & 32 \\
Learning-rate schedule & OneCycleLR (10\% warmup + cosine annealing) \\
Early stopping & Patience of 20 epochs on validation AUC \\
Loss function & Weighted BCE ($\mathrm{pos\_weight} = n_{\mathrm{neg}} / n_{\mathrm{pos}}$) \\
Training timesteps & $\{200, 400, 600\}$ (pooled) \\
Projection dimension $d$ & 256 \\
Number of attention heads & 4 \\
\bottomrule
\end{tabular}
}
\end{table}

\vspace{0.5em}
\noindent\textbf{Layer selection.}
We train separate physics verifiers for candidate layers
($\ell \in \{5, 10, 15, 20, 25\}$ for CogVideoX-2B; up to $\ell = 35$ for
the deeper CogVideoX-5B and Wan backbones) and pick the layer with the highest
validation AUC.
The causal verifier consistently outperforms the flat linear probe of the main
paper, which we attribute to its causal temporal modeling.
The selected layers and resulting AUCs are the ones reported in the main
paper's verifier comparison table.

\vspace{0.5em}
\noindent\textbf{Reward-gradient guidance.}
Guidance uses the same verifier as selection.
At each guidance step we recompute the truncated forward pass through blocks
$1,\dots,\ell_{\mathrm{best}}$ with gradients enabled, evaluate
$\log r_t$, and backpropagate to the latent; the resulting gradient is
normalized by a per-frame norm clip of 1.0 and injected with strength
$w_{\mathrm{phy}} = 0.5$ inside the window $t \in [400, 600]$, i.e.\ only
between the two selection checkpoints.
Everything outside the truncated subgraph stays under
\texttt{torch.no\_grad}.

\vspace{0.5em}
\noindent\textbf{DPO baseline.}
The Diffusion-DPO baseline trains LoRA adapters (rank 64, $\alpha$ 64) on the
attention projections of the frozen backbone with the standard Diffusion-DPO
objective ($\beta = 2500$), using 2{,}000 same-caption preference pairs built
from the VideoPhy physics annotations, AdamW at learning rate $10^{-5}$,
gradient accumulation 2, and 1{,}000 optimizer steps.
The frozen reference model is obtained by disabling the adapters, so no second
copy of the backbone is required.

\vspace{0.5em}
\noindent\textbf{LikePhys protocol.}
The training-free likelihood baseline scores a video by the negative denoising
error $-\,\mathbb{E}_t \lVert \hat{v} - v^{*} \rVert^2$ at
$t \in \{200, 400, 600\}$ under caption conditioning, with noise draws
shared across videos (paired comparison) and four draws per timestep.
The noising and target follow each backbone's native parameterization
(v-prediction for CogVideoX, flow matching for Wan).
AUC is computed against the same annotations as the trained verifiers.

\subsection{Feature Extraction Pipeline}
\label{sec:supp_features}

\noindent\textbf{DiT feature dimensions.}
CogVideoX (both 2B and 5B) processes video latents as a flattened
sequence of spatio-temporal tokens. For 49-frame videos at
$480 \times 720$ resolution, the VAE produces latent tensors of shape
$[13, C, 60, 90]$, corresponding to 13 temporal frames with spatial
resolution $60 \times 90$. The DiT then patchifies and flattens these
latents into $13 \times 30 \times 45 = 17{,}550$ video tokens, with 226
text-conditioning tokens prepended to the sequence. At each layer, we
extract the full hidden state
$\mathbf{h}^{(\ell)} \in \mathbb{R}^{(226 + 17{,}550) \times D}$,
discard the first 226 text tokens, reshape the remaining video tokens
into $[13, 30 \times 45, D]$, and mean-pool over the spatial dimension
to obtain $\mathbf{f}^{(\ell)} \in \mathbb{R}^{13 \times D}$.
For Wan~2.1-14B, text conditioning is applied exclusively through
cross-attention, so the hidden states contain only video tokens. The
extraction pipeline is therefore simpler: we reshape the video-token
sequence from $[B, T \cdot H \cdot W, D]$ to $[B, T, H \cdot W, D]$ and
then mean-pool over the spatial dimension. No text-token removal is
required.

\vspace{0.5em}
\noindent\textbf{Hook mechanism.}
Features are extracted via a PyTorch forward hook registered on the
target transformer block. For CogVideoX, the hook is attached to
\texttt{transformer.transformer\_blocks[$\ell$]}, whereas for Wan it is
attached to \texttt{transformer.blocks[$\ell$]} due to differences in
module naming. The hook records the \emph{output} of the transformer
block after the residual connection, rather than intermediate attention
or MLP activations. Because these features are already computed during
the standard forward pass, the additional computational overhead is
negligible.

\vspace{0.5em}
\noindent\textbf{Text conditioning.}
CogVideoX adopts a hybrid text-conditioning scheme. Specifically, text
information is injected via cross-attention in every transformer block,
while text tokens are also concatenated with video tokens in the input
sequence. During feature extraction, we provide the text prompt and
therefore obtain text-conditioned features, but discard the explicit
text tokens from the extracted hidden states, retaining only the
video-token representations. As a result, textual information is
reflected implicitly through the conditioned video features rather than
through explicit text-token features.
By contrast, Wan~2.1-14B performs text conditioning exclusively through
cross-attention, and text tokens are never concatenated with the
video-token sequence. This removes the need for text-token stripping
and leads to a simpler feature extraction pipeline.

\subsection{Video Generation Configuration}
\label{sec:supp_generation}

\noindent\textbf{Backbones.}
All backbones generate 49-frame videos at $480 \times 720$ with
classifier-free guidance scale 6.0.
The CogVideoX backbones use the DDIM scheduler and Wan~2.1-14B uses
UniPCMultistep.

\vspace{0.5em}
\noindent\textbf{Trajectory selection.}
For progressive trajectory selection, we initialize $N = 4$ parallel
trajectories using independent random seeds (defined by the base seed,
prompt index, and trajectory index), and perform selection at
checkpoints $t \in \{600, 400\}$ with a keep ratio of $\rho = 0.5$. As
a result, the number of active trajectories decreases from
$4 \rightarrow 2 \rightarrow 1$ across the two selection stages. This
configuration is used consistently across all three backbones.
Checkpoints are not placed at $t{=}800$ because discrimination at high
noise levels is unreliable, nor at $t{=}200$ because late selection in
the schedule yields minimal computational savings.

\vspace{0.5em}
\noindent\textbf{Scoring at checkpoints.}
At each checkpoint, we perform one additional forward pass for each active
trajectory using only the conditional (positive) prompt embedding, without
classifier-free guidance, and the physics verifier scores the captured
layer-$\ell$ features.
With $N{=}4$ and two checkpoints this amounts to six extra passes on top of
the ${\approx}120$-pass selection schedule (${\approx}5\%$), already included
in all reported wall-clock times; the verifier head itself adds less than
0.01\,s per trajectory.

\vspace{0.5em}
\noindent\textbf{VAE decoding.}
Only the final surviving trajectory is decoded by the VAE. We enable
VAE tiling but disable slicing, as we observed that enabling both
simultaneously can produce solid-color artifacts in the 3D causal VAEs
used by both CogVideoX and Wan.

\begin{table}[t]
\centering
\caption{Summary of the video generation configuration for the three
backbones evaluated in this work.}
\label{tab:gen_config}
\small
\setlength{\tabcolsep}{2pt}
\resizebox{\linewidth}{!}{%
\begin{tabular}{@{}lccc@{}}
\toprule
\textbf{Parameter} & \textbf{CogVideoX-2B} & \textbf{CogVideoX-5B} & \textbf{Wan~2.1-14B} \\
\midrule
Resolution & $480 \times 720$ & $480 \times 720$ & $480 \times 720$ \\
Number of frames & 49 (13 latent) & 49 (13 latent) & 49 (13 latent) \\
Sampling scheduler & DDIM & DDIM & UniPCMultistep \\
Guidance scale & 6.0 & 6.0 & 6.0 \\
\midrule
Number of trajectories ($N$) & 4 & 4 & 4 \\
Selection checkpoints ($\mathcal{C}$) & $\{600, 400\}$ & $\{600, 400\}$ & $\{600, 400\}$ \\
Keep ratio ($\rho$) & 0.5 & 0.5 & 0.5 \\
Verifier layer & 10 & 10 & 10 \\
\midrule
Primary hardware & A5000 (24\,GB) & L40S (48\,GB) & H200 (80\,GB) \\
\bottomrule
\end{tabular}
}
\end{table}

\vspace{0.5em}
\noindent\textbf{CogVideoX-5B specific notes.}
CogVideoX-5B requires explicit rotary positional embeddings
(\texttt{image\_rotary\_emb}) computed using a 2D grid that matches the
spatial latent dimensions. These embeddings must be provided at every
transformer forward pass; omitting them leads to all-black outputs. By
contrast, CogVideoX-2B uses learned positional embeddings and does not
require this additional input.

\vspace{0.5em}
\noindent\textbf{Wan~2.1-14B specific notes.}
Wan~2.1-14B (approximately 28\,GB in bf16) fits on a single 80\,GB GPU
(H200). We observe that distributing Wan across multiple devices can
introduce tiling and blurring artifacts.

\vspace{0.5em}
\noindent\textbf{Reproducibility.}
For prompt $i$, the first trajectory is initialized with seed $42 + i$,
and subsequent trajectories use seeds $42 + i + 1$, $42 + i + 2$, and
so on. This setup ensures that trajectory~0 in our method shares the
same initial noise as the single-seed baseline, enabling a controlled
comparison of the effect of trajectory selection.

\begin{table}[t]
\centering
\caption{Wall-clock time breakdown per video across backbones and
modes.}
\label{tab:timing}
\setlength{\tabcolsep}{2pt}
\resizebox{\linewidth}{!}{%
\begin{tabular}{@{}llcccc@{}}
\toprule
\textbf{Backbone (GPU)} & \textbf{Component} & \textbf{Baseline} & \textbf{Rand.\ Sel.} & \textbf{Best-of-4} & \textbf{Ours} \\
\midrule
\multirow{4}{*}{\shortstack[l]{CogVideoX-2B\\(A5000 24\,GB)}}
  & Sampling             & 183s & 454s & 732s & 448s \\
  & Scoring              & ---  & ---  & 3s   & 3s   \\
  & Decode + offload     & 60s  & 60s  & 60s  & 60s  \\
  & \textbf{Total}       & \textbf{204s} & \textbf{477s} & \textbf{778s} & \textbf{490s} \\
\midrule
\multirow{3}{*}{\shortstack[l]{CogVideoX-5B\\(L40S 48\,GB)}}
  & Sampling             & 137s & ---  & ---  & 357s \\
  & Decode + offload     & 12s  & ---  & ---  & 12s  \\
  & \textbf{Total}       & \textbf{149s} & ---  & ---  & \textbf{374s} \\
\midrule
\multirow{3}{*}{\shortstack[l]{Wan 2.1-14B\\(H200 80\,GB)}}
  & Sampling             & 117s$^\dagger$ & --- & --- & 391s \\
  & Decode               & 1s   & ---  & ---  & 1s   \\
  & \textbf{Total}       & \textbf{118s}$^\dagger$ & --- & --- & \textbf{393s} \\
\bottomrule
\multicolumn{6}{@{}l}{\footnotesize $^\dagger$Estimated on H200. Measured baseline on 2$\times$L40S is ${\sim}$305s due to cross-GPU overhead.}
\end{tabular}%
}
\end{table}

\vspace{0.5em}
\noindent\textbf{Hardware and timing.}
\Cref{tab:timing} provides wall-clock time breakdowns per video across
all three backbones, measured from SLURM job logs.
For CogVideoX-5B, the per-step cost is ${\sim}$4.6s per trajectory on
L40S. Under the $4 {\to} 2 {\to} 1$ schedule, sampling splits into a
four-trajectory phase (${\sim}$220s), a two-trajectory phase
(${\sim}$55s), and a single-trajectory phase (${\sim}$55s), plus six
scoring forwards (${\sim}$27s).
For Wan~2.1-14B on H200, the per-step cost is ${\sim}$3.9s per
trajectory despite $5.6\times$ more parameters than CogVideoX-2B,
because batch-1 inference is memory-bandwidth bound and H200 provides
${\sim}$5.5$\times$ higher bandwidth (4,800\,GB/s HBM3e \vs\
864\,GB/s on L40S). The two selection checkpoints reduce
per-step cost from ${\sim}$15.5s (4 trajectories) to ${\sim}$8s (2) to
${\sim}$5.5s (1).

\subsection{Physics-IQ Evaluation Protocol}
\label{sec:supp_physicsiq}

Physics-IQ conditions the model on the switch frame of a real recorded scene
and scores the generated continuation against the actual future.
We generate with CogVideoX-5B-I2V, conditioning on the
official switch frames and scene descriptions, and evaluate with the official
toolbox (binary masks, spatial/spatiotemporal IoU, weighted spatial IoU, MSE).
Generated clips are resampled to the benchmark's 30\,FPS, 5-second format
before scoring.
We cover 42 of the 66 scenarios, each from all three camera perspectives (252
videos per method); the remaining scenarios were not generated for both
methods under our compute budget, and the evaluator's completeness checks are
relaxed accordingly, scoring only complete scenario/perspective sets.

\section{Additional Experimental Analysis}
\label{sec:supp_experiments}


\subsection{Comparison with VideoREPA}
\label{sec:supp_videorepa}

VideoREPA~\cite{zhang2025videorepa} is a training-based alternative that
fine-tunes CogVideoX-5B by aligning token-relation structure in DiT features
with a self-supervised video encoder (VideoMAEv2), and thus complements the
training-based DPO baseline of the main paper: both spend training compute to
move the backbone's weights, whereas our mechanisms spend inference compute on
a frozen backbone.
We evaluate the officially released VideoREPA-5B adapter with its official
inference configuration (LoRA fused at scale $\alpha/r = 64/128$), generating
all 160 PhyGenBench prompts under one shared protocol (seed 42, single
trajectory) and scoring with the same PhyGenEval pipeline.

\begin{table}[t]
\centering
\setlength{\tabcolsep}{4pt}
\resizebox{\linewidth}{!}{%
\begin{tabular}{@{}lccccc@{}}
\toprule
Method & Final & S1 & S2 & S3 & Time \\
\midrule
VideoREPA                   & 0.492 & \textbf{1.87} & \textbf{0.97} & 1.63 & $1\times$ \\
DPO (LoRA)                  & 0.475 & 1.74 & 0.78 & 1.68 & $1\times$ \\
$+$ Selection $+$ Reward Gradient (Ours) & \textbf{0.496} & 1.79 & 0.95 & \textbf{1.71} & $2.0\times$ \\
\bottomrule
\end{tabular}%
}
\caption{\textbf{VideoREPA comparison} on PhyGenBench (CogVideoX-5B, all
methods measured by us under one shared protocol, seed 42, and scored with
the same PhyGenEval pipeline). Time is relative single-prompt inference
cost.}
\label{tab:supp_videorepa}
\end{table}

VideoREPA is the strongest training-based reference we measure (0.492),
essentially matching our inference-time result (0.496): it leads on
single-frame semantics and event ordering but trails on naturalness
(S3 1.63 \vs\ our 1.71).
The two are also not exclusive, since our mechanisms operate on frozen
weights and could be applied on top of VideoREPA's fine-tuned backbone; we
leave this combination to future work.
Notably, the two prompts on which our base model collapses to degenerate
near-constant videos also collapse under VideoREPA's fine-tuned weights,
suggesting these failure modes are rooted in the base model rather than
addressable by light physics-oriented fine-tuning.

\subsection{Score Distribution Analysis}
\label{sec:supp_scores}

\begin{figure}[t]
\centering
\begin{subfigure}[t]{0.48\linewidth}
  \includegraphics[width=\linewidth]{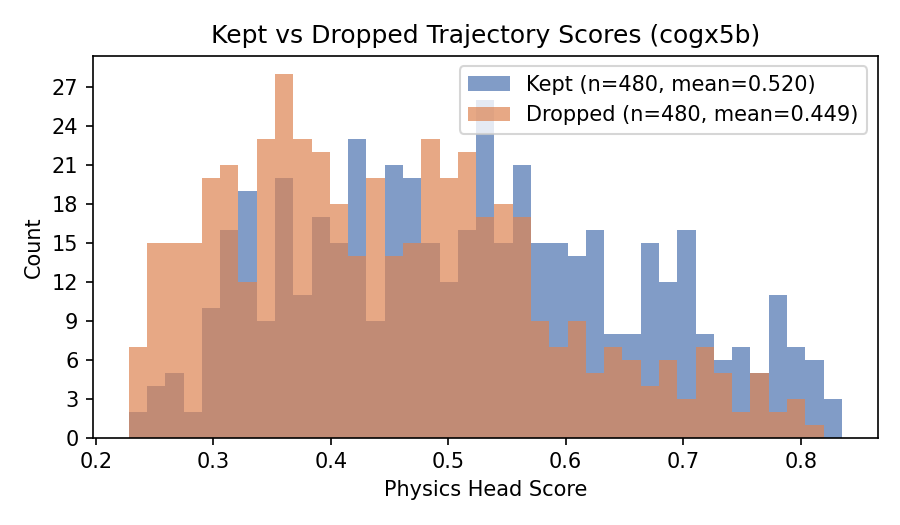}
  \caption{CogVideoX-5B: kept \vs\ dropped}
\end{subfigure}\hfill
\begin{subfigure}[t]{0.48\linewidth}
  \includegraphics[width=\linewidth]{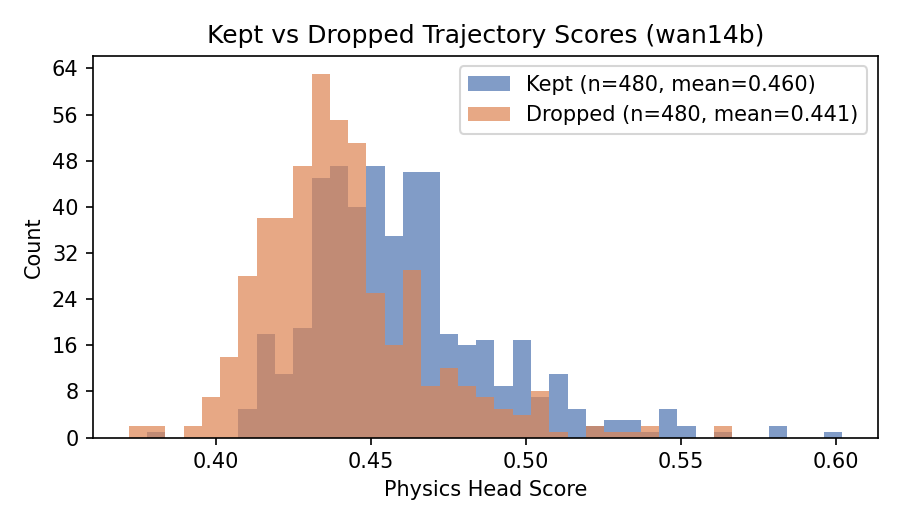}
  \caption{Wan 2.1-14B: kept \vs\ dropped}
\end{subfigure}\\[4pt]
\begin{subfigure}[t]{0.48\linewidth}
  \includegraphics[width=\linewidth]{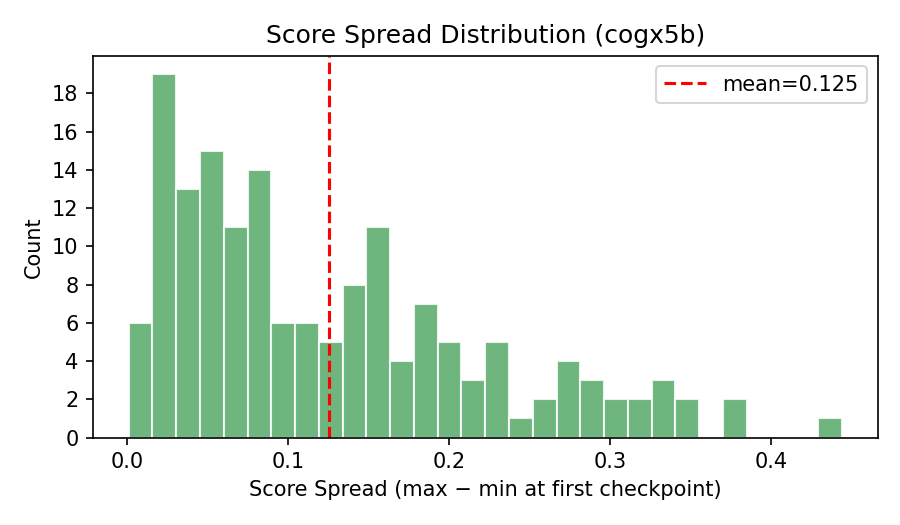}
  \caption{CogVideoX-5B: score spread}
\end{subfigure}\hfill
\begin{subfigure}[t]{0.48\linewidth}
  \includegraphics[width=\linewidth]{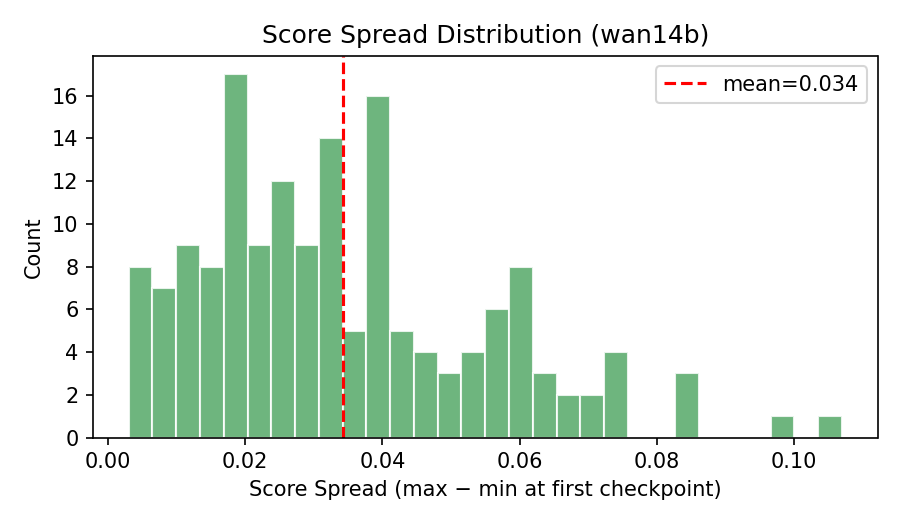}
  \caption{Wan 2.1-14B: score spread}
\end{subfigure}
\caption{Verifier score analysis. Top: kept \vs\ dropped trajectory
scores at checkpoints. Bottom: per-prompt score spread (max $-$ min) at
the first checkpoint. 5B shows meaningful separation; Wan shows
near-complete overlap.}
\label{fig:score_analysis}
\end{figure}

\Cref{fig:score_analysis} compares the verifier's scoring behavior on
CogVideoX-5B and Wan~2.1-14B, complementing the cross-backbone results
reported in the main paper. On 5B, kept and dropped trajectories show
visible separation ($\Delta\text{mean}{=}0.071$) and per-prompt score
spread averages 0.125, indicating the head provides a meaningful
selection signal. On Wan, the two distributions nearly completely
overlap ($\Delta\text{mean}{=}0.019$) and score spread averages only
0.034 ($3.7\times$ smaller), confirming that the Wan verifier operates
near chance and the PhyGenBench improvement (+0.043) is attributable to
multi-trajectory diversity rather than physics-informed guidance.

\begin{figure}[t]
\centering
\includegraphics[width=\linewidth,height=0.35\textheight,keepaspectratio]{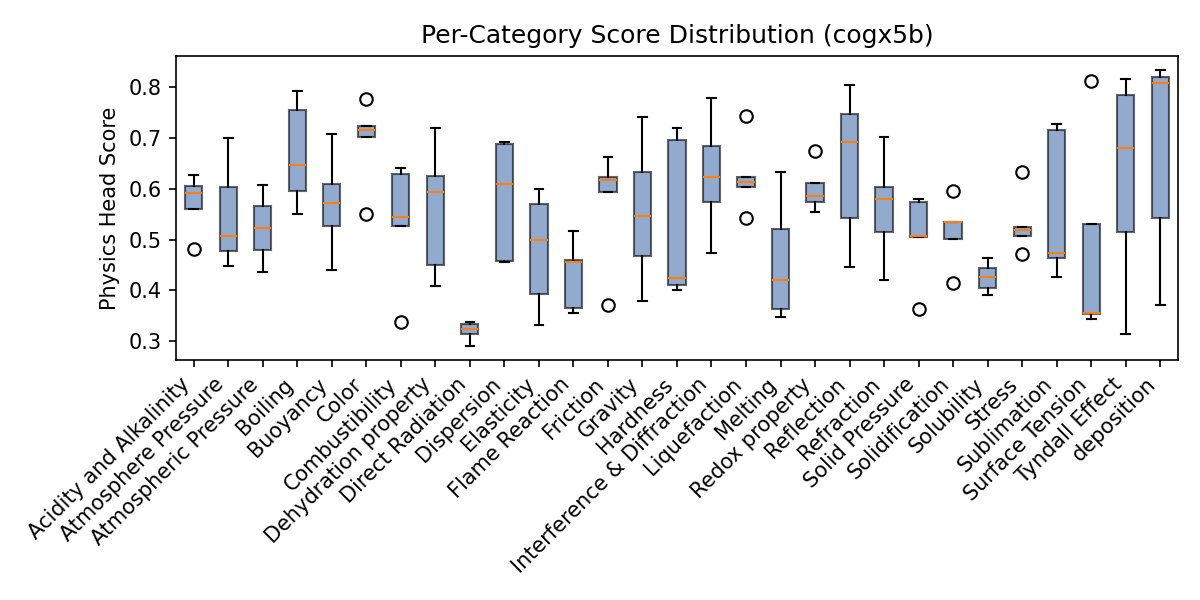}\\[4pt]
\includegraphics[width=\linewidth,height=0.35\textheight,keepaspectratio]{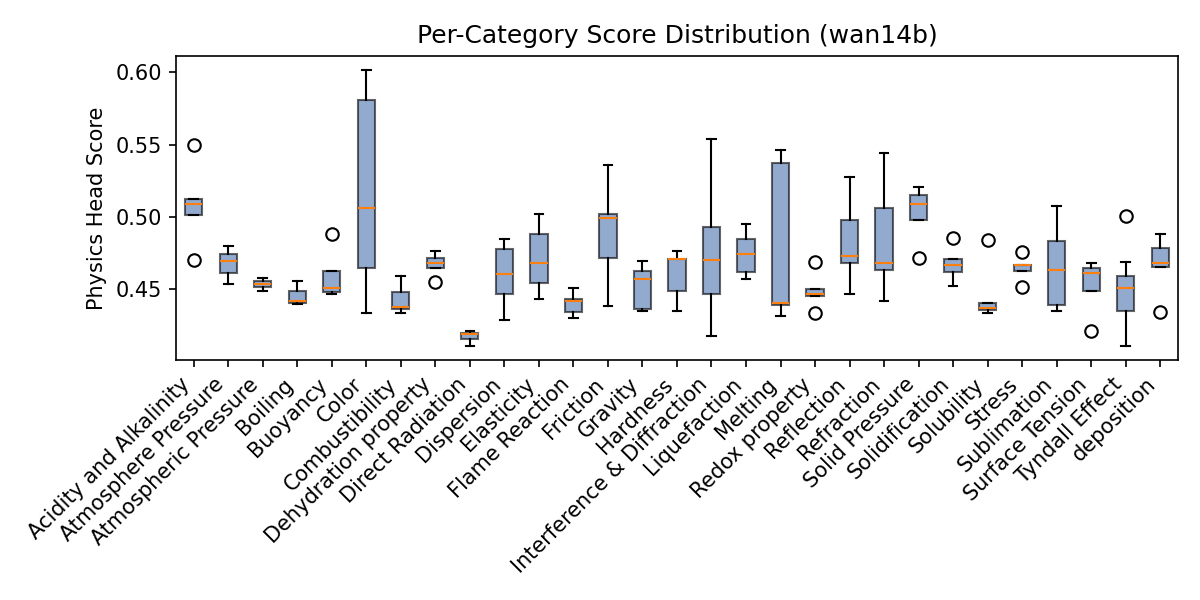}
\caption{Per-category score distributions for CogVideoX-5B (top) and
Wan~2.1-14B (bottom). On 5B, temporal phenomena (Color, Deposition,
Boiling) score highest while spatial/material-specific phenomena
(Direct Radiation, Flame Reaction) score lowest. On Wan, categories
cluster in $[0.42, 0.52]$.}
\label{fig:per_category}
\end{figure}

\paragraph{Per-category analysis.}
\Cref{fig:per_category} breaks down verifier scores by the 27 physical
law categories in PhyGenBench. On CogVideoX-5B, the verifier exhibits
meaningful category-level specialization. It achieves the highest
confidence on phenomena characterized by salient temporal state
changes: Color mixing (0.693), Deposition (0.675), Boiling (0.668),
Tyndall Effect (0.635), and Reflection (0.649), indicating that the
verifier has learned to detect progressive physical dynamics such as
phase transitions, light scattering, and color blending from
intermediate DiT features. Categories with lower scores, such as Direct
Radiation (0.320) and Flame Reaction (0.430), typically involve precise
spatial reasoning or domain-specific material knowledge (\eg, shadow
geometry, characteristic flame colors), which are underrepresented in
the VideoPhy training distribution and suggest a clear avenue for
improvement through targeted data augmentation. Notably, this
category-dependent structure itself is an informative signal: it
demonstrates that the verifier captures physically meaningful
distinctions rather than relying on superficial quality cues. On
Wan~2.1-14B, category-level variation is substantially reduced, with
all means falling within $[0.42, 0.52]$, consistent with the distribution
mismatch discussed above, and motivating backbone-matched training data
as the primary path toward stronger cross-backbone transfer.

\section{More Qualitative Results}
\label{sec:visualization}

\begin{figure}[t]
\centering
\setlength{\tabcolsep}{1pt}
\small
\newcommand{\catlab}[1]{\rotatebox[origin=c]{90}{\scriptsize\textsf{#1}}}
\begin{tabular}{@{}r@{\hspace{3pt}} c @{\hspace{6pt}} c@{}}
 & \textbf{Baseline} & \textbf{Ours} \\[2pt]
\raisebox{0.5\height}{\catlab{Mechanics}} &
\includegraphics[width=0.475\linewidth]{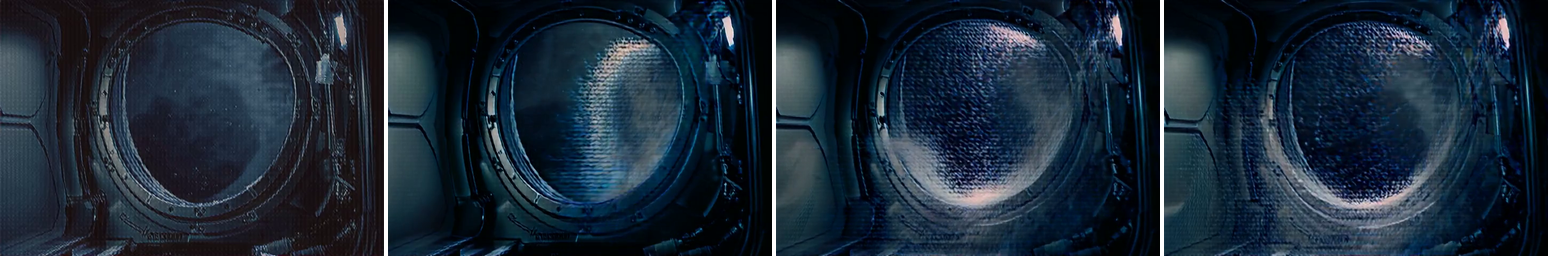} &
\includegraphics[width=0.475\linewidth]{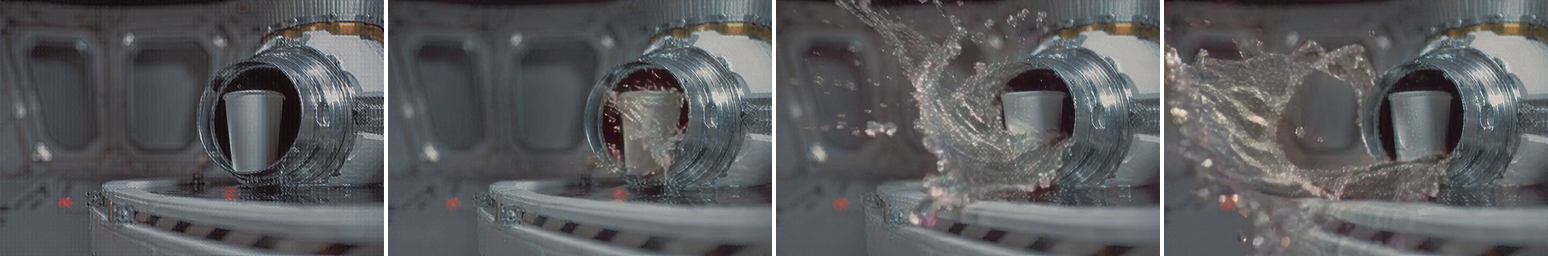} \\[-2pt]
& \multicolumn{2}{c}{\parbox{0.92\linewidth}{\centering\scriptsize\textit{%
``A cup of water is slowly poured out in the space station, releasing the liquid into the surrounding area''}}} \\[4pt]
& \multicolumn{2}{c}{\parbox{0.92\linewidth}{\scriptsize%
Baseline pours water downward under terrestrial gravity. Ours forms a floating liquid mass consistent with microgravity, where liquid disperses in multiple directions.}} \\[6pt]
\raisebox{0.5\height}{\catlab{Mechanics}} &
\includegraphics[width=0.475\linewidth]{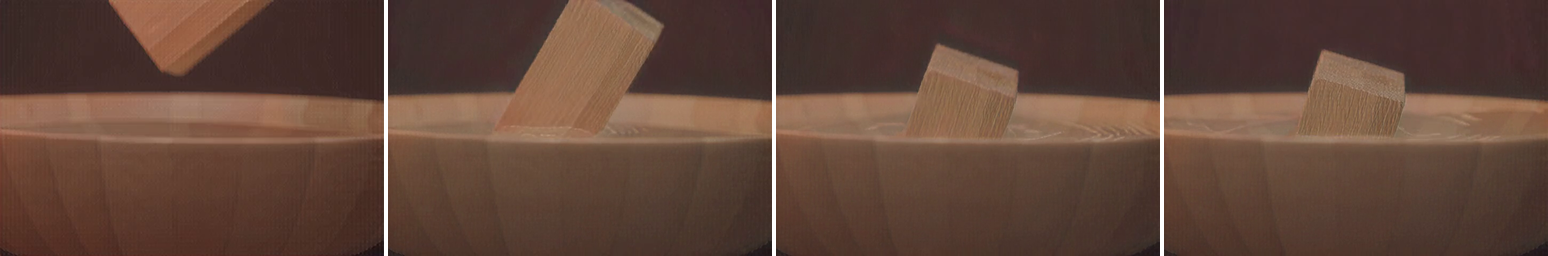} &
\includegraphics[width=0.475\linewidth]{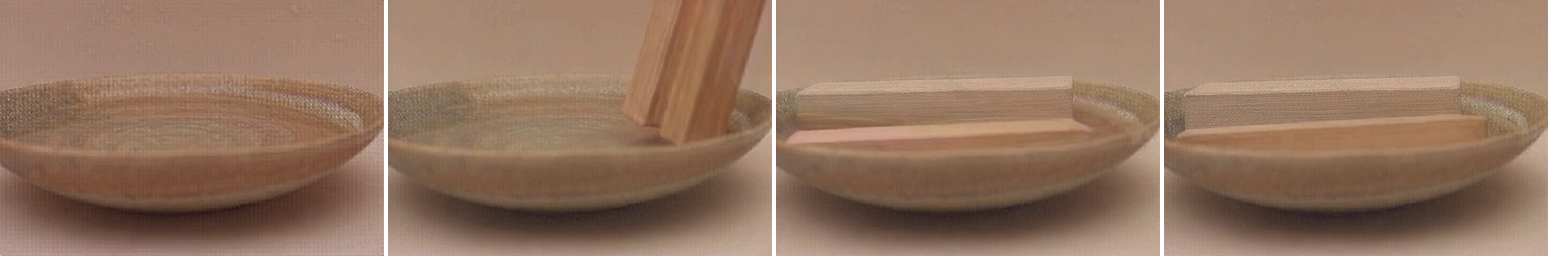} \\[-2pt]
& \multicolumn{2}{c}{\parbox{0.92\linewidth}{\centering\scriptsize\textit{%
``A piece of wood block is gently placed on the surface of a bowl filled with water''}}} \\[4pt]
& \multicolumn{2}{c}{\parbox{0.92\linewidth}{\scriptsize%
Wood is less dense than water and should float. Baseline shows ambiguous contact with the water surface. Ours depicts the wood block resting on the water, consistent with buoyancy.}} \\[6pt]
\raisebox{0.5\height}{\catlab{Optics}} &
\includegraphics[width=0.475\linewidth]{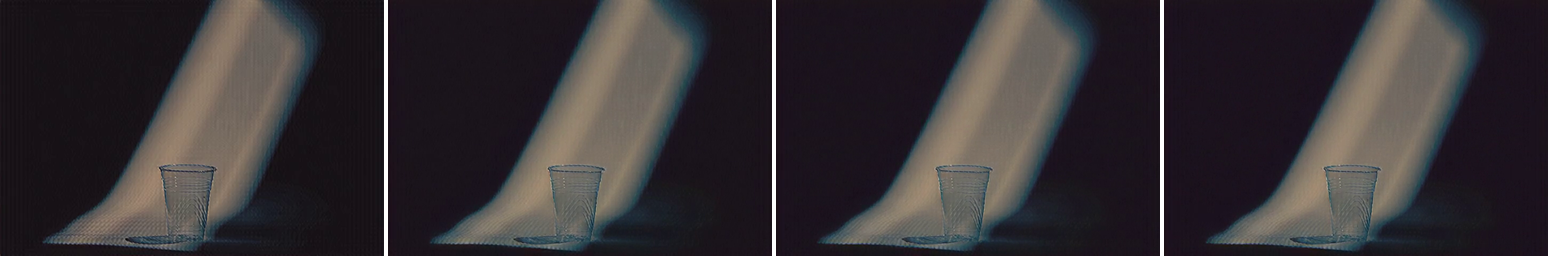} &
\includegraphics[width=0.475\linewidth]{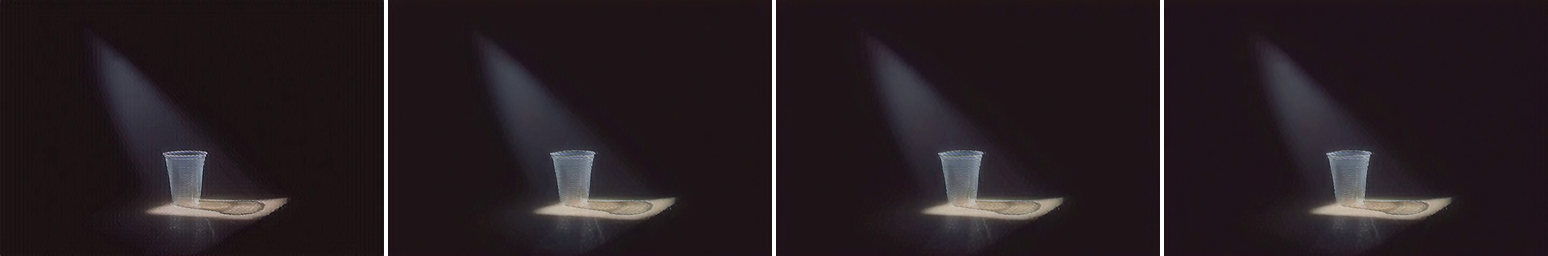} \\[-2pt]
& \multicolumn{2}{c}{\parbox{0.92\linewidth}{\centering\scriptsize\textit{%
``A ray of light is shining diagonally on a plastic cup in the dark, with the shadow of the plastic cup appearing at the bottom''}}} \\[3pt]
& \multicolumn{2}{c}{\parbox{0.92\linewidth}{\scriptsize%
Baseline shows diffuse illumination without a clear shadow. Ours produces a well-defined diagonal light beam with the cup's shadow visible at the bottom.}} \\[6pt]
\raisebox{0.5\height}{\catlab{Optics}} &
\includegraphics[width=0.475\linewidth]{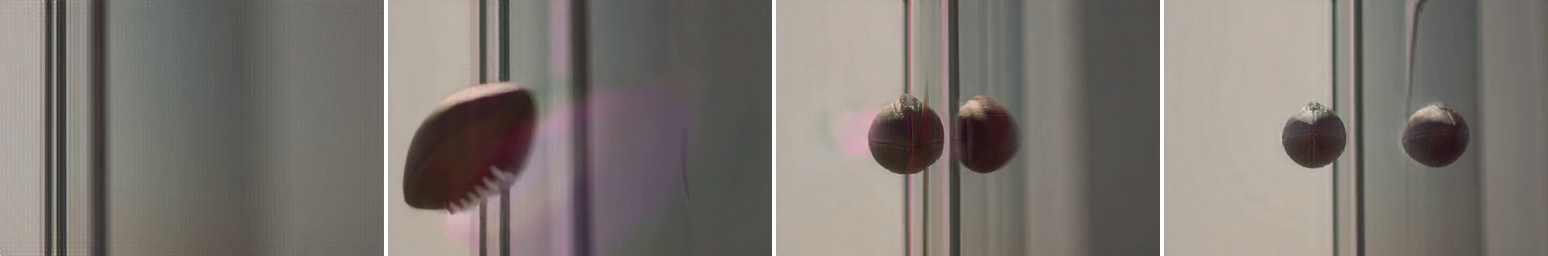} &
\includegraphics[width=0.475\linewidth]{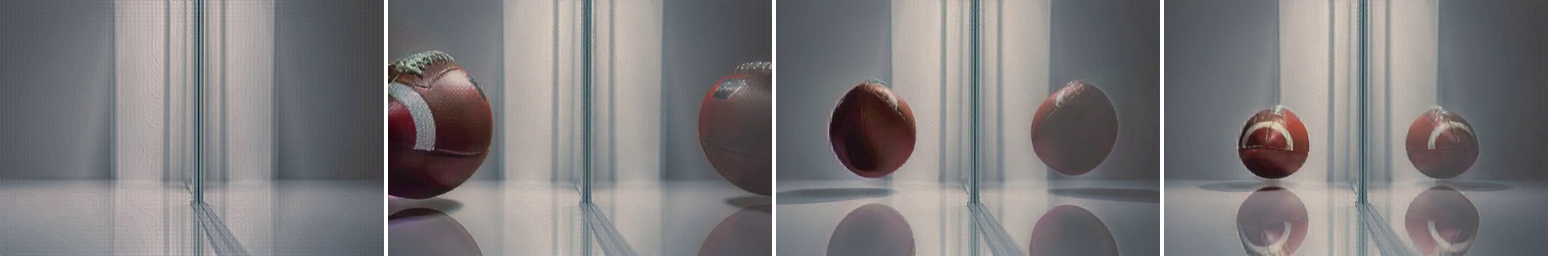} \\[-2pt]
& \multicolumn{2}{c}{\parbox{0.92\linewidth}{\centering\scriptsize\textit{%
``A football is rolling towards a vertical window, with the reflection in the window moving closer synchronously''}}} \\[3pt]
& \multicolumn{2}{c}{\parbox{0.92\linewidth}{\scriptsize%
Baseline shows a blurry football mid-air with no coherent reflection in the window (e.g., Frame 2). Ours renders the football on a reflective surface near glass.}} \\[6pt]
\raisebox{0.5\height}{\catlab{Thermal}} &
\includegraphics[width=0.475\linewidth]{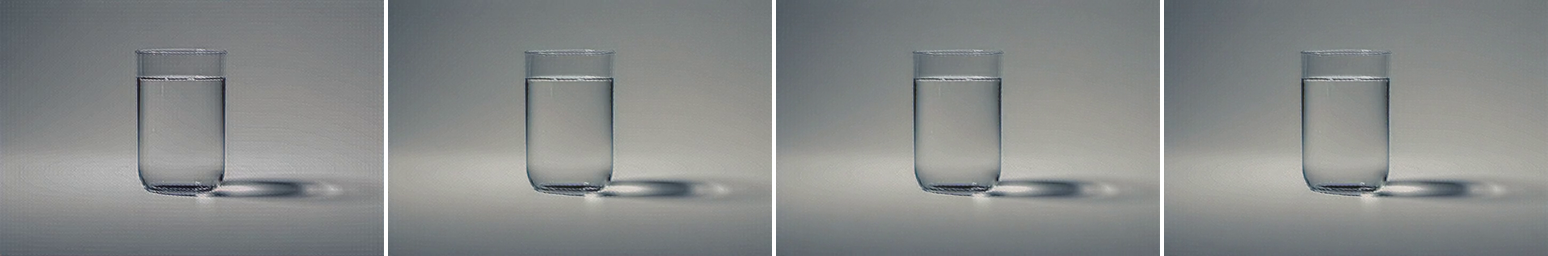} &
\includegraphics[width=0.475\linewidth]{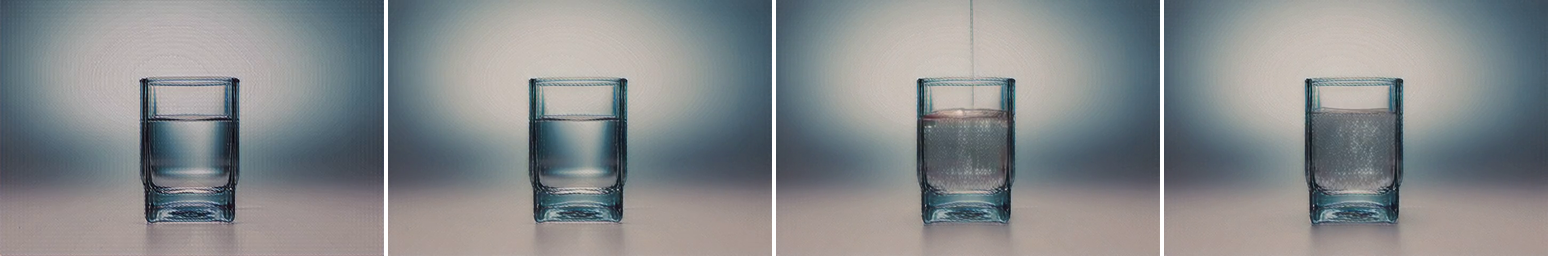} \\[-2pt]
& \multicolumn{2}{c}{\parbox{0.92\linewidth}{\centering\scriptsize\textit{%
``A timelapse captures the transformation of a glass of water as the temperature significantly drops to very low levels''}}} \\[3pt]
& \multicolumn{2}{c}{\parbox{0.92\linewidth}{\scriptsize%
Baseline remains clear liquid throughout. Ours shows progressive opacity change, consistent with solidification.}} \\[6pt]
\raisebox{0.5\height}{\catlab{Thermal}} &
\includegraphics[width=0.475\linewidth]{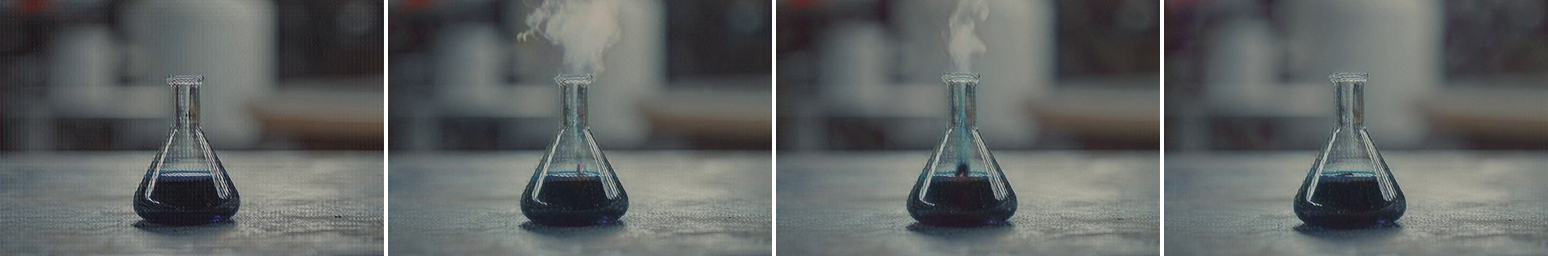} &
\includegraphics[width=0.475\linewidth]{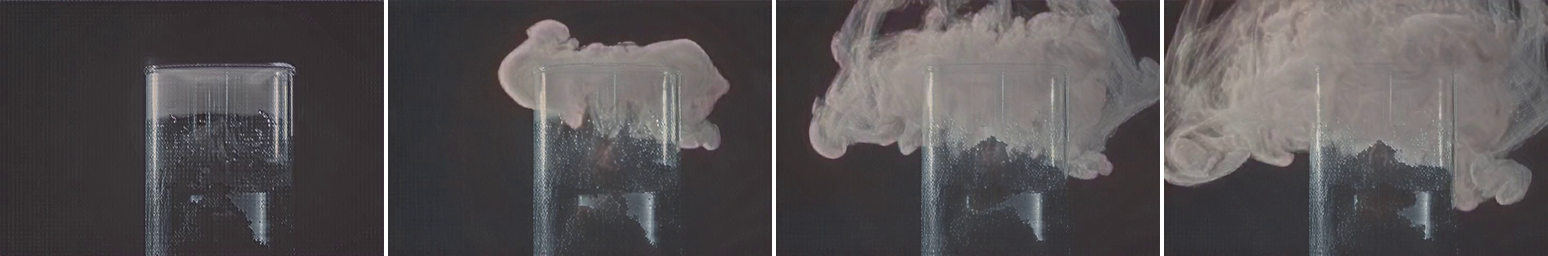} \\[-2pt]
& \multicolumn{2}{c}{\parbox{0.92\linewidth}{\centering\scriptsize\textit{%
``A timelapse captures the transformation of arsenic trioxide as it is exposed to gradually increasing temperature at room temperature''}}} \\[3pt]
& \multicolumn{2}{c}{\parbox{0.92\linewidth}{\scriptsize%
Baseline depicts a dark liquid in a flask with minor steam. Ours shows a white solid with expanding vapor, consistent with sublimation.}} \\[6pt]
\raisebox{0.5\height}{\catlab{Material}} &
\includegraphics[width=0.475\linewidth]{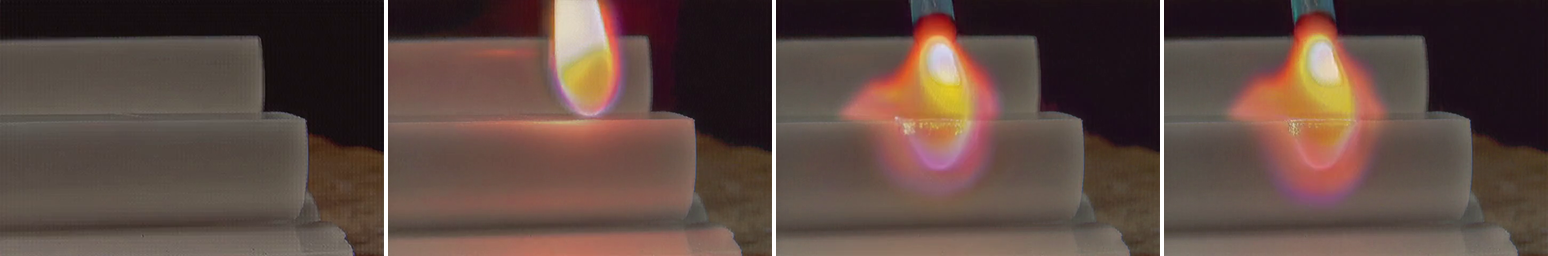} &
\includegraphics[width=0.475\linewidth]{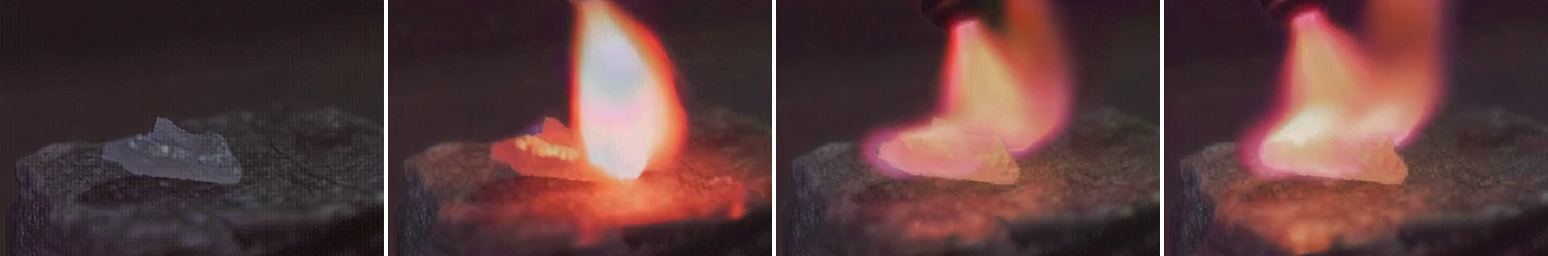} \\[-2pt]
& \multicolumn{2}{c}{\parbox{0.92\linewidth}{\centering\scriptsize\textit{%
``A piece of lithium is ignited, emitting a vivid and unique flame as it burns steadily''}}} \\[3pt]
& \multicolumn{2}{c}{\parbox{0.92\linewidth}{\scriptsize%
Lithium burns with a crimson flame. Baseline shows a yellow-cored one.}} \\
\end{tabular}
\vspace{-2mm}
\caption{\textbf{Qualitative comparison} on PhyGenBench prompts
(Wan-14B, 4 uniformly sampled frames per video). Each row shows
baseline (left) \vs\ guided (right), the generation prompt, and a brief
physical analysis.}
\label{fig:qualitative_wan}
\end{figure}

\Cref{fig:qualitative_wan} presents additional Wan~2.1-14B comparisons
across mechanics, optics, thermal, and material prompts, each with a
brief physical analysis of where the baseline violates the underlying
law and how the guided result differs.

\section{Failure Cases}
\label{sec:supp_failure}

\begin{figure}[t]
\centering
\small
\begin{minipage}{\linewidth}\centering
\begin{minipage}[t]{0.04\linewidth}\phantom{X}\end{minipage}%
\hfill
\begin{minipage}[t]{0.465\linewidth}\centering\textbf{Baseline}\end{minipage}%
\hfill
\begin{minipage}[t]{0.465\linewidth}\centering\textbf{Ours}\end{minipage}\\[4pt]
\noindent
\begin{minipage}[t]{0.04\linewidth}\vspace{0pt}\centering\rotatebox[origin=c]{90}{\scriptsize\textsf{Material}}\end{minipage}%
\hfill
\begin{minipage}[t]{0.465\linewidth}\vspace{0pt}\includegraphics[width=\linewidth]{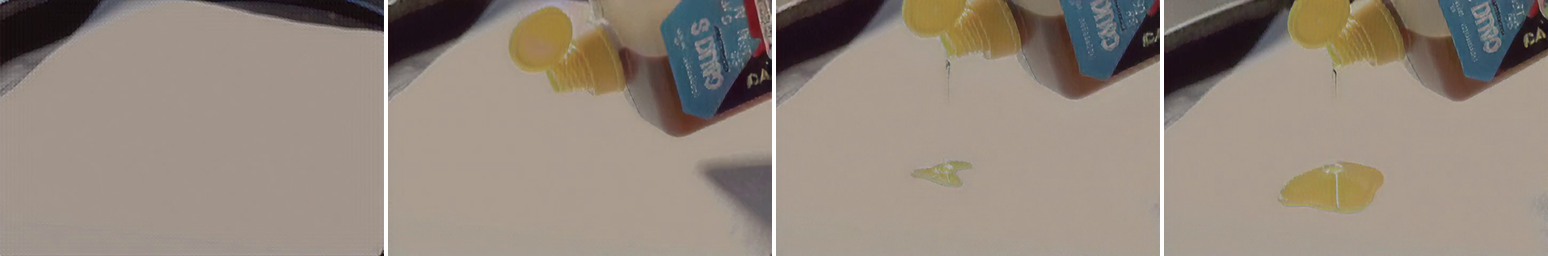}\end{minipage}%
\hfill
\begin{minipage}[t]{0.465\linewidth}\vspace{0pt}\includegraphics[width=\linewidth]{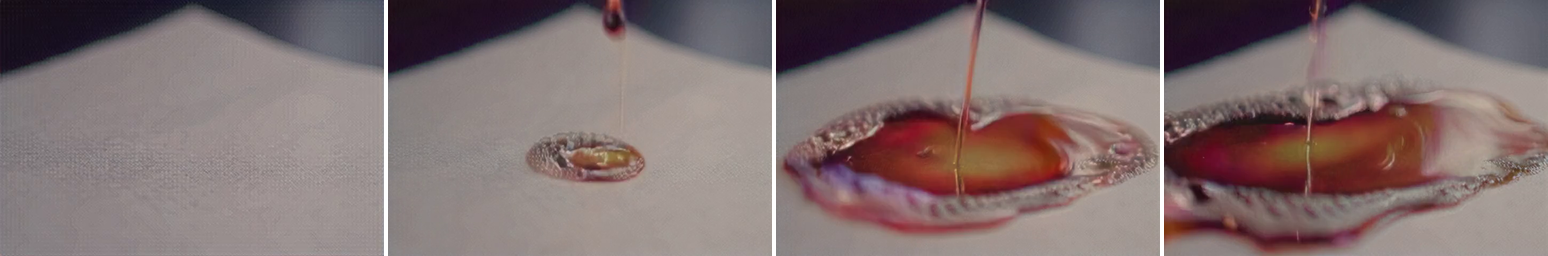}\end{minipage}\\[4pt]
\begin{minipage}{\linewidth}\centering\scriptsize\textit{``A timelapse captures the reaction as concentrated sulfuric acid is poured onto a sheet of paper''}\end{minipage}\\[4pt]
\begin{minipage}{\linewidth}\scriptsize Sulfuric acid is colorless and causes black carbonization. Baseline shows a yellow liquid with no reaction; ours shows a red liquid with paper degradation. Neither depicts the correct acid color or carbonization.\end{minipage}\\[12pt]
\noindent
\begin{minipage}[t]{0.04\linewidth}\vspace{0pt}\centering\rotatebox[origin=c]{90}{\scriptsize\textsf{Material}}\end{minipage}%
\hfill
\begin{minipage}[t]{0.465\linewidth}\vspace{0pt}\includegraphics[width=\linewidth]{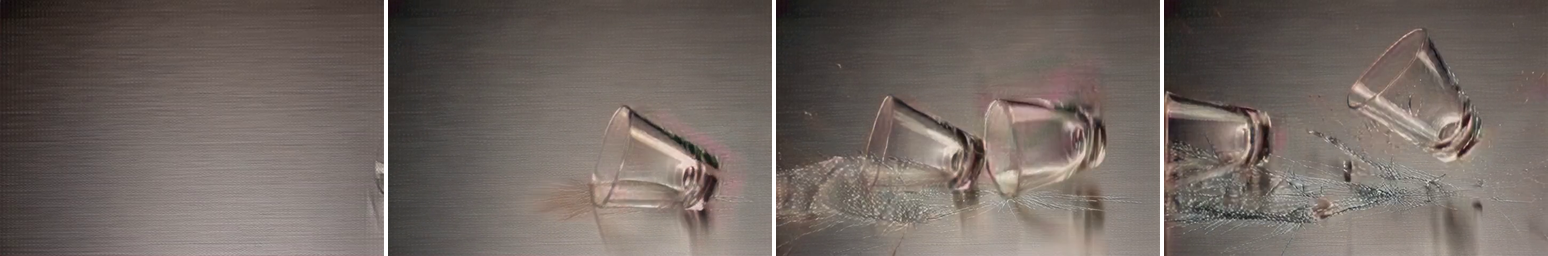}\end{minipage}%
\hfill
\begin{minipage}[t]{0.465\linewidth}\vspace{0pt}\includegraphics[width=\linewidth]{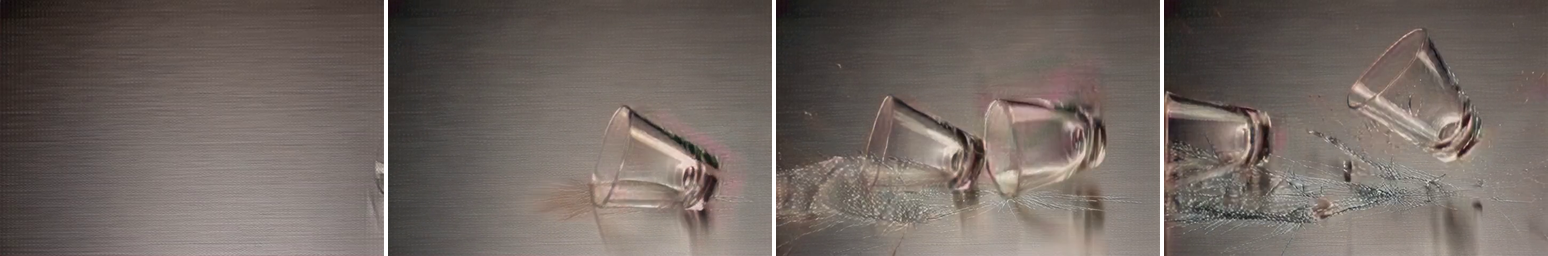}\end{minipage}\\[4pt]
\begin{minipage}{\linewidth}\centering\scriptsize\textit{``A flimsy, brittle glass cup is propelled with force towards a solid, metallic surface, where it collides upon impact''}\end{minipage}\\[4pt]
\begin{minipage}{\linewidth}\scriptsize The cup should shatter into fragments. Both produce nearly identical results: the cup cracks but never exhibits realistic brittle fracture.\end{minipage}
\end{minipage}
\caption{\textbf{Failure cases} (Wan-14B). Both rows show cases where
all candidate trajectories are physically incorrect, leaving the
verifier unable to select a better outcome.}
\label{fig:failure}
\end{figure}

\Cref{fig:failure} illustrates representative failure cases. These
concentrate in material properties and chemical reaction prompts, which
require domain-specific knowledge, such as characteristic flame
colors, acid appearances, and fracture mechanics, that goes beyond the
common-sense physical reasoning captured by our training data.
These phenomena also involve dramatic transformations in material form
(carbonization, shattering, phase changes), which current video
diffusion models struggle to synthesize faithfully. When no candidate
trajectory is physically correct, selection reaches a fundamental
ceiling.

\end{document}